\documentclass[10pt,conference]{ieeetran}
\usepackage{graphicx}

\usepackage{cathywumath}
\usepackage{amsmath}
\usepackage{hyperref}
\usepackage{tikz}
\usepackage{aircraftshapes}
\usepackage[short,12hr]{datetime}
\usepackage{url}
\usetikzlibrary{math,decorations.markings,arrows,shapes.geometric,patterns} 
\usepackage{pgfplots}
\usetikzlibrary{intersections, pgfplots.fillbetween}
\usepackage{wrapfig}
\usepackage{hyphenat}
\usepackage{times}
\usepackage{cuted}

\newif\ifdraft
\draftfalse

\newcommand{\nishant}[1]{\ifdraft{\color{orange}[{\bf Nishant}: #1]}\fi}

\newcommand{\eat}[1]{}

\begin{document}
\title{Automating Geometric Proofs of Collision Avoidance with Active Corners}
%
%
%
\author{
\IEEEauthorblockN{Nishant Kheterpal\IEEEauthorrefmark{1}} \IEEEauthorblockA{\textit{University of Michigan}} 
\and
\IEEEauthorblockN{Elanor Tang} \IEEEauthorblockA{\textit{University of Michigan}} 
\and
\IEEEauthorblockN{Jean-Baptiste Jeannin} \IEEEauthorblockA{\textit{University of Michigan}}
}

%
%
%


%
\maketitle              
\thispagestyle{plain}
\pagestyle{plain}
\begingroup\renewcommand\thefootnote{*}
\footnotetext{Emails: \texttt{\{nskh, elanor, jeannin\}@umich.edu}}
\endgroup

%


\begin{abstract}

Avoiding collisions between obstacles and vehicles such as cars, robots or aircraft is essential to the development of automation and autonomy.
To simplify the problem, many collision avoidance algorithms and proofs consider vehicles to be a point mass, though the actual vehicles are not points.
In this paper, we consider a convex polygonal vehicle with nonzero area traveling along a 2-dimensional trajectory.
We derive an easily-checkable, quantifier-free formula to check whether a given obstacle will collide with the vehicle moving on the planned trajectory.
We apply our method to two case studies of aircraft collision avoidance and study its performance.



\end{abstract}


\section{Introduction}

Preventing collisions with obstacles or foreign objects is crucial when developing autonomous capabilities for robots, cars, aircraft, and many other vehicles.
As such, collision avoidance remains a major research theme of the autonomy, robotics, and formal methods communities.
In particular, for safety-critical tasks such as vehicles interacting with humans or animals, it is imperative to provide \emph{formal} proofs that the vehicle will not collide with agents in its environment.

In many papers studying trajectory planning or collision avoidance, e.g. \cite{mitsch2013provably,althoff2021setpropagation-survey,girard2005reachability}, the vehicle is modeled as a point, and the volume -- or surface area -- occupied by the vehicle is ignored.
In reality, land and air vehicles are not points but have a certain volume, and contact of any external object with any part of the vehicle would constitute a collision.
In this paper, we present a novel, automated, and general technique to \emph{transform} a planned trajectory of a vehicle with volume  
into explicit boundaries of the region in which an obstacle will not be at risk of a collision.
This transformation provides an efficient, runtime-checkable test to determine whether a given obstacle will collide with a vehicle on the planned trajectory, even when the vehicle has volume.

Given a part of a trajectory $\TT$, a vehicle occupying the volume $v(x_\TT,y_\TT)$ when centered on position $(x_\TT,y_\TT)$ along the trajectory, and a point-obstacle 
$(x_O,y_O)$, the vehicle will not collide with the obstacle if and only if:
\begin{align}
\forall(x_\TT,y_\TT)\in\TT, (x_O, y_O)\not\in v(x_\TT,y_\TT)
\label{eqn:implicit}
\end{align}
In the rest of the paper, we will call this formulation the \emph{implicit} formulation of collision avoidance. This implicit formulation is a correct definition, but it has one major drawback: because of the universal quantifier on $(x_\TT,y_\TT)$, it is not easy to check systematically or at runtime whether an obstacle is indeed at risk of a collision.
Ideally, we would want to obtain a quantifier-free, easily checkable formula that is equivalent to (\ref{eqn:implicit}); in the rest of this paper we will call that formula, which represents a region in the plane, the \emph{explicit} formulation.
In theory, one could use quantifier elimination, but for trajectories containing more than a few symbolic parameters, the algorithm does not finish in a reasonable time due to its doubly-exponential time complexity in the number of variables \cite{QEDoublyExpDavenport}.

This issue arose before, notably in the verification of the Next-Generation Aircraft Collision Avoidance System ACAS~X~\cite{jeannin2015acastacas,JeanninAcas17}. In that work, the formal proof of correctness was divided into: (i) establishing the trajectory of the aircraft from its equations of motion, leading to a formula of the form of (\ref{eqn:implicit}); and (ii) establishing an equivalent quantifier-free formula that can be checked efficiently at runtime.
Both tasks required a proof in the KeYmaera~X theorem prover, with significant manual effort \cite{DBLP:journals/jar/Platzer17/KeymaeraXToolPaper}. 
The object of this paper is to automate and generalize task (ii) of this process.

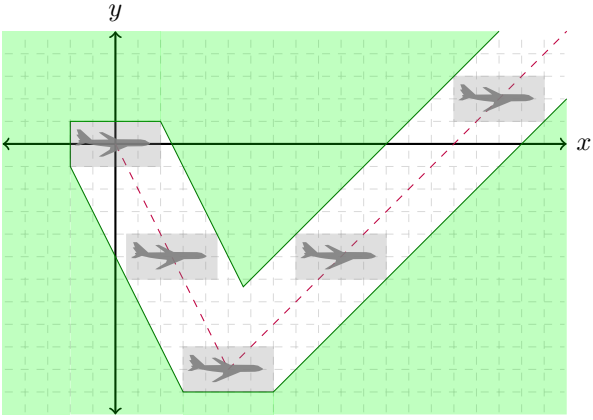
\begin{figure}
    \centering
    \begin{tikzpicture}[scale=0.3]
        \draw[help lines, color=gray!30, dashed] (\xmin + 0.1, \ymin + 0.1) grid (\xmax - 0.1, \ymax - 0.1);
        \draw[<->,thick] (\xmin,0)--(\xmax,0) node[right]{$x$};
        \draw[<->,thick] (0,\ymin)--(0,\ymax) node[above]{$y$};
        
        \draw [purple, dashed] (0, 0) -- (\vertexx, \vertexy); 
        \draw [purple, dashed] (\vertexx, \vertexy) -- (20, 5); 
        
        \draw [fill, gray, nearly transparent] (-\w, -\h) rectangle (\w, \h); 
        \node [aircraft side,fill=gray,fill opacity=0.9,minimum width=1cm] at (0 + 0.2, 0) {};
        \draw [fill, gray, nearly transparent] (-\w + 2.5, -\h - 5) rectangle (\w + 2.5, \h - 5);
        \node [aircraft side,fill=gray,fill opacity=0.9,minimum width=1cm] at (2.7, -5) {};
        \draw [fill, gray, nearly transparent] (-\w + 5, -\h - 10) rectangle (\w + 5, \h - 10); 
        \node [aircraft side,fill=gray,fill opacity=0.9,minimum width=1cm] at (5.2, -10) {};
        \draw [fill, gray, nearly transparent] (-\w + 10, -\h - 5) rectangle (\w + 10, \h - 5); 
        \node [aircraft side,fill=gray,fill opacity=0.9,minimum width=1cm] at (10 + 0.2, -5) {};
        \draw [fill, gray, nearly transparent] (-\w + 17, -\h + 2) rectangle (\w + 17, \h + 2); 
        \node [aircraft side,fill=gray,fill opacity=0.9,minimum width=1cm] at (17 + 0.2, 2) {};
        
        \draw[black!60!green] (17, 5 + \h + \w - 3) -- (17/3, -19/3) -- (\w, \h) -- (-\w, \h) -- (-\w, -\h) -- (\vertexx-\w, \vertexy-\h) -- (\vertexx+\w, \vertexy-\h) -- (20, 5 - \h - \w);

        \fill[green,nearly transparent] (-\w, \ymin) -- (-\w, \ymax) -- (\xmin, \ymax) -- (\xmin, \ymin);
        \fill[green,nearly transparent] (-\w, -\h) -- (-\w, \ymin) -- (\vertexx-\w, \ymin) -- (\vertexx-\w, \vertexy-\h) -- cycle;
        \fill[green, nearly transparent] (\vertexx-\w, \ymin) -- (\vertexx-\w, \vertexy-\h) -- (\vertexx+\w, \vertexy-\h) -- (\vertexx+\w, \ymin) -- cycle;
        \fill[green, nearly transparent] (\vertexx+\w, \vertexy-\h) -- (\vertexx+\w, \ymin) -- (\xmax, \ymin) -- (20, 5 - \h - \w) -- cycle;
        
        \fill[green,nearly transparent] (-\w, \h) -- (-\w, \ymax) -- (\w, \ymax) -- (\w, \h);
        \fill[green, nearly transparent] (\w, \ymax) -- (\w, \h) -- (17/3, -19/3) -- (17, 5 + \h + \w - 3); 
        
    \end{tikzpicture}
    \caption{Safe region for a rectangle with $w=2, h=1$ and its center following Equation (\ref{eq:nominal-planar}). The trajectory is dashed in purple, safe region  shaded in green, and unsafe region is unshaded.}
    \label{fig:safe-region-example}
\end{figure}

In order to automate task (ii), we propose a different approach based on geometric intuition.
Let us examine an example of a rectangular vehicle performing a simple maneuver (Figure~\ref{fig:safe-region-example}). The central idea of the method presented in this paper is that the boundaries of the explicit formulation are either \emph{trajectories of a corner} of the vehicle or \emph{sides of the vehicle} at a few particular points. 
The use of corners is geometrically intuitive, yet we are not aware of past work that uses corners in this way to prove reachability properties.
\nishant{should we get rid of this last sentence?}

The corners to consider at every point depend on the slope of the trajectory: for a rectangular vehicle, the boundaries follow the top-right and bottom-left corners when the vehicle's velocity is directed ``northwest" (towards the top left) or southeast; and, symmetrically, the boundaries follow the top-left and bottom-right corners when the vehicle's velocity is directed towards the northeast or southwest of the plane.
We call these corners \emph{active corners}.
But this is not enough: at points where the trajectory switches from following one set of corners to another, the boundary may follow a side of the vehicle at that point, e.g., its bottom boundary at the lowest point of the trajectory on Figure~\ref{fig:safe-region-example}.
We call these points \emph{transition points}.
By capturing the motion of these boundaries -- both corners depending on the slope or sides of the vehicle -- we can construct a quantifier-free formula equivalent to (\ref{eqn:implicit}), corresponding to its equivalent explicit formulation.
Our approach is fully symbolic with no approximation.

In this paper, we formalize and generalize our approach to different trajectories and polygons and show how to find active corners and transitions symbolically and how to form a quantifier free explicit formulation equivalent to the input implicit formulation. We carefully prove that our transformation is both sound (that any obstacle shown safe using the explicit formulation is safe using the implicit formulation) and complete (no obstacles that are actually safe appear unsafe using the explicit formulation). Finally, we detail a fully symbolic Python implementation of our work and present an evaluation of its performance on two applications from previous papers (where we fully automate results that had required significant manual proof effort) and a third non-polynomial example.

\section{Overview}
\label{sec:overview}

\tikzmath{\xmin = -5; \xmax = 20; \ymin = -12; \ymax = 5; \w = 2; \h = 1; \vertexx = 5; \vertexy = -10;}

This section provides an overview of our approach, by walking through the different steps of a simple example constructing a geometric safe region used to verify obstacle avoidance of aircraft. At present, our method applies only to two-dimensional planar motion due to the increased complexity of three-dimensional motion when analyzing trajectories and polyhedra. The example uses linear motion and a rectangle, though the method generalizes to other planar motion and convex polygons, as detailed in Section \ref{proof-section}.

\tikzmath{\x1 = 0; \y1 =0; 
\x2 = \x1 + 3; \y2 =\y1 +2; } 


\subsection{Trajectory}
Consider the planar side-view of an airplane flying, initially descending at constant velocity and then ascending with constant velocity. For simplicity, assume the aircraft has infinite acceleration. In this example, we represent the bounds of the aircraft as an axis-aligned rectangle with width $2w$ and height $2h$ that moves in the $(x, y)$ plane. The airplane begins at the origin and moves in the plane with piecewise trajectory $\mathcal{T}$: 
\begin{equation}\mathcal{T} = \begin{cases}
y = -2x & x \in [0, 5] \\ 
y = x - 15 & x \in [5, \infty)
\end{cases}
\label{eq:nominal-planar}
\end{equation}

\eat{
\begin{figure}
    \centering
    \includegraphics[width=0.7\linewidth]{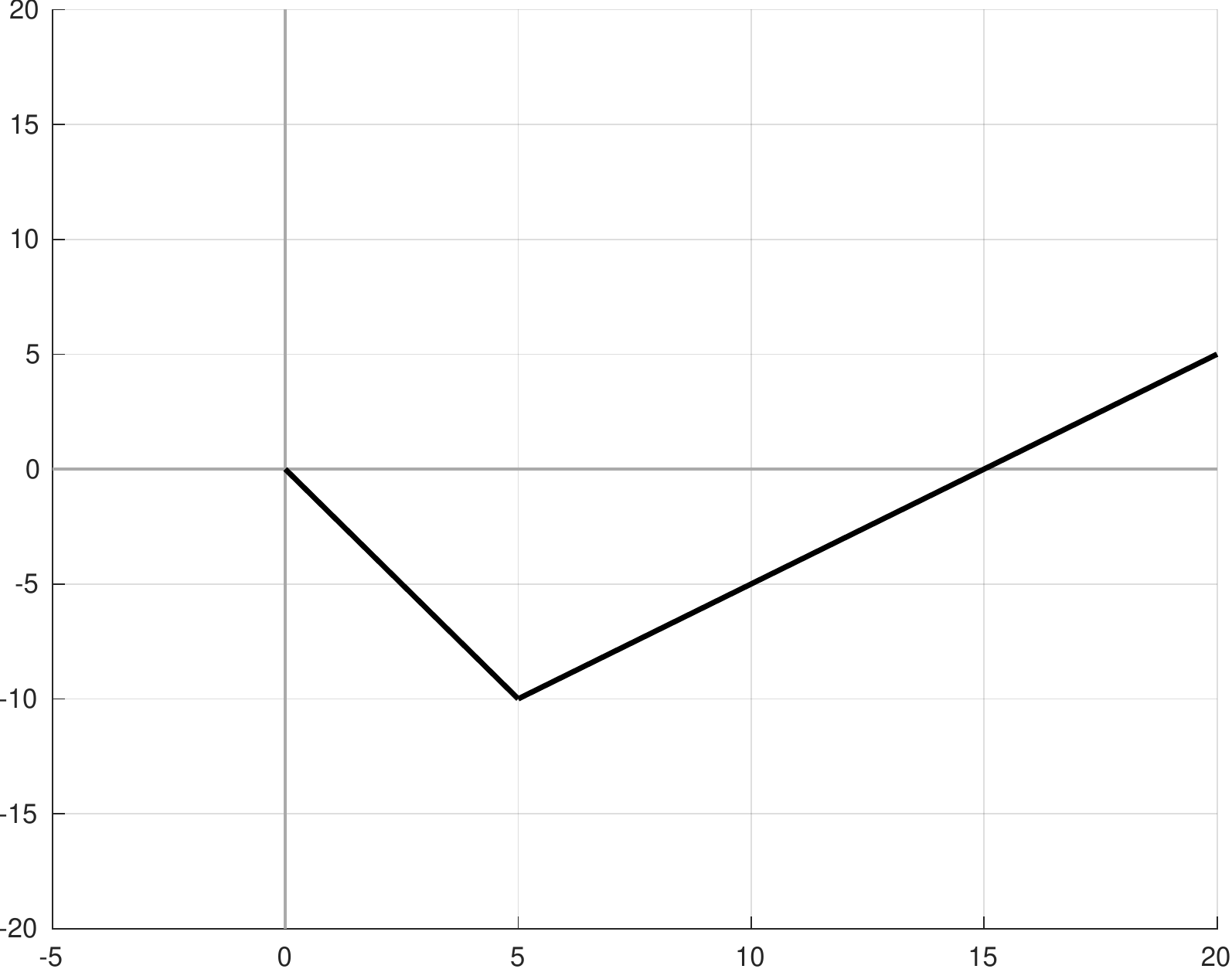}
    \caption{Trajectory of rectangle center following Equation (\ref{eq:nominal-planar}).}
    \label{fig:trajectory}
\end{figure}
}

\subsection{Implicit Formulation}

Suppose the rectangle translates with its center moving along this piecewise trajectory. Additionally, assume there is a point obstacle at $(x_O, y_O)$ to be avoided; that is, the rectangle never intersects the obstacle. Then we can state an quantified (or \textit{implicit}) formulation of obstacle avoidance:
\begin{equation}
    \forall(x_\mathcal{T}, y_\mathcal{T}) \in \mathcal{T}, \left(|x_O - x_\mathcal{T}| > w \vee |y_O - y_\mathcal{T}| > h\right)
    \label{eq:example-implicit-safety-property}
\end{equation}
The implicit formulation of the safe region (\ref{eq:example-implicit-safety-property}) straightforwardly represents a safety property of obstacle avoidance - if an object moving with its center fixed along the nominal trajectory $\mathcal{T}$ is far enough away (either width $w$ in the $x$-axis or height $h$ in the $y$-axis) from a point obstacle at $(x_O, y_O)$, it is safe. Here we use \textit{safe region} to mean the set of all obstacle locations for which collision is avoided.



\subsection{Explicit Formulation}
The need for a quantifier-free equivalent of \eqref{example-implicit-safety-property} motivates an \textit{explicit formulation} of the safe region for the obstacle.
The goal of this work is to automate the generation of such a formulation. 
We compute the reachable set of the object as it moves along a trajectory in order to compute the complement of the safe region: the \textit{unsafe region}.
We can express the \textit{unsafe region} (the set of all locations for which an obstacle will collide with the object as it moves along a given trajectory) directly as a union of regions in the plane, each defined (in this case) by an intersection of linear inequalities bounding the region, plotted in Figure~\ref{fig:safe-region-example}. The inequalities are presented in Appendix \ref{fig-1-inequalities}.

\section{Algorithm}
\label{sec:algorithm}





\subsection{Preliminaries} 
\label{sec:definitions}
\eat{ 
We consider trajectories in the $xy$-plane with the form $f(x, y) = 0$, where $f$ can be any function with real output. 
}

In this work, we define the \textit{safe region} as the set of obstacle locations where, given a polygon's trajectory, a collision will not occur. Correspondingly, the \textit{unsafe region} is called that because it will be unsafe if an obstacle invades that area. 
As such, the \textit{unsafe region} corresponds to the reachable set of the polygon as it moves along a trajectory. We can define a quantified representation of the safe region: the \textit{implicit formulation} from (\ref{eqn:implicit}).
We also use the term \textit{explicit formulation} in this work; we use that to mean an equivalent to the \textit{implicit formulation}, but without quantifiers like $\forall$ and $\exists$.
Our method applies to convex polygons; concave polygons and circles are not within the scope of this work. In this paper, we discuss polygons with central symmetry for ease of exposition, though the method straightforwardly extends to irregular and asymmetric convex polygons (Appendix~\ref{sec:extensions}). 

We consider two-dimensional planar trajectories defined piecewise, with each piece a function $y=f(x)$ or $x=f(y)$ and $f$ a $C^1$ function (differentiable and having a continuous derivative). Trajectories must have a \textit{finite} number of these $C^1$ pieces. The pieces themselves need not be continuous, though the applications we study do include continuous piecewise trajectories.
The subdomains for the piecewise trajectory must be non-overlapping and exhaustive, meaning their union should cover the entire domain of the trajectory. Polygons move along the trajectory without rotating.
Since the polygons translate along the trajectory, there is a constant vector offset $\begin{bmatrix}\Delta x_i \\ \Delta y_i\end{bmatrix}$ from the center to the $i$-th vertex, and there are $n$ vertices for an $n$-sided polygon. 
Thus, the trajectory for vertex $i$ is $y - \Delta y_i = f(x - \Delta x_i)$ or $x - \Delta x_i = f(y - \Delta y_i)$. 
We consider the trajectories of all vertices of the polygon in an attempt to bound its motion and compute the reachable set of the object as it moves along the trajectory.


\eat{
For trajectories of the form $f(x,y) = 0$, the slopes of the tangent to the trajectory will be \begin{equation}
    \frac{\partial y}{\partial x} = \frac{\partial f / \partial x}{\partial f / \partial y}
    \label{eq:implicit_deriv}
\end{equation}
evaluated at $(x,y)$ points on the trajectory. Therefore, \begin{equation}
    \theta = \arctan \left(\frac{\partial f / \partial x}{\partial f / \partial y}\right)
    \label{eq:theta}
\end{equation}
}

\subsection{Active Corners}
\label{sec:active-corners}

\renewcommand\deg{{}^\circ}
Throughout this section we consider a trajectory of the form $y = f(x)$; the case for $x=f(y)$ is symmetric.
The boundaries of the safe region are typically formed by the trajectories of a pair of opposite corners of the vehicle (Figure~\ref{fig:hexagon-with-angles}) -- we call this pair of corners \emph{active corners}. \nishant{Rephrase to avoid ``typically" here - long version is that 1) corners may not be exactly opposite for asymmetric polygons and 2) the boundaries also consist of the polygon at transition points} We choose the active corners to represent the outermost extent of the object along the trajectory; as such, their motion bounds the safe region. Which corners are active depends on the slope of the trajectory (which can be computed from the derivative of $f$) and the shape of the convex object. A corner $v_i$ is active when the slope $\theta$ of the trajectory is between the slopes of the sides adjacent to $v_i$; when a corner is active, its opposite corner is also active based on the symmetry of the polygon. 
More precisely, if we number the corners $v_1$ through $v_n$ counter-clockwise (with $v_{n+1}=v_0$ and $v_{-1}=v_n$), corner $v_i$ is active if and only if the slope $\theta$ of the trajectory is in the angle interval $[\angle\overrightarrow{v_{i-1}v_i},\angle\overrightarrow{v_i v_{i+1}}]$, or symmetrically in the angle interval $[\angle\overrightarrow{v_{i+1}v_i},\angle\overrightarrow{v_i v_{i-1}}]$. Because the direction of the trajectory is inconsequential for our purpose, $\theta$ is modulo $180\deg$.

For example, on the hexagon of Figure~\ref{fig:hexagon-with-angles}, $v_1$ and $v_4$ are active when $\theta\in[0\deg,60\deg]\cup[180\deg,240\deg]$; $v_2$ and $v_5$ are active when $\theta\in[60\deg,120\deg]\cup[240\deg,300\deg]$; and $v_3$ and $v_6$ are active when $\theta\in[120\deg,180\deg]\cup[300\deg,360\deg]$.
At \text{transition points} (where the active corners change), the boundary of the safe region may not follow an active corner. We detail what happens then in Section~\ref{sec:notches}.

\tikzmath{\hexshiftx=1.1;\hexshifty=1.8;}
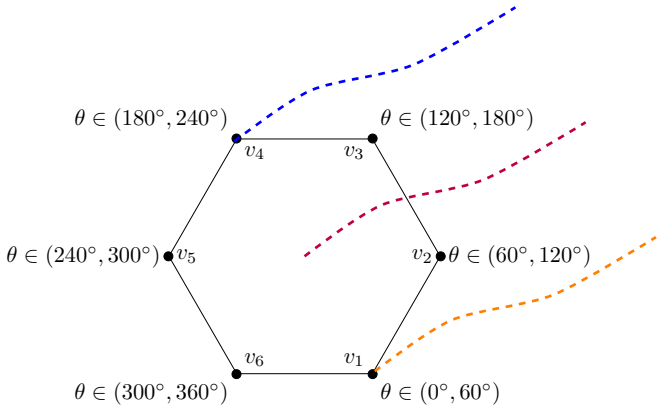
\begin{figure}
    \centering
    \resizebox{\columnwidth}{!}{%
        \begin{tikzpicture}
            \path
            node[
              regular polygon,
              regular polygon sides=6,
              draw,
              inner sep=1.3cm,
            ] (hexagon) {}
            %
            (hexagon.corner 1) node[above right] {$\theta \in (120\deg, 180\deg)$}
            (hexagon.corner 2) node[above left] {$\theta \in (180\deg, 240\deg)$}
            (hexagon.corner 3) node[left] {$\theta \in (240\deg, 300\deg)$}
            (hexagon.corner 4) node[below left] {$\theta \in (300\deg, 360\deg)$}
            (hexagon.corner 5) node[below right] {$\theta \in (0\deg, 60\deg)$}
            (hexagon.corner 6) node[right] {$\theta \in (60\deg, 120\deg)$}
            
            (hexagon.corner 1) node[below left] {$v_3$}
            (hexagon.corner 2) node[below right] {$v_4$}
            (hexagon.corner 3) node[right] {$v_5$}
            (hexagon.corner 4) node[above right] {$v_6$}
            (hexagon.corner 5) node[above left ] {$v_1$}
            (hexagon.corner 6) node[left] {$v_2$}
            
            plot[
              mark=*,
              samples at={1, ..., 6},
            ] (hexagon.corner \x)
            ;
            
            \draw[purple, dashed, very thick] plot [smooth] coordinates { (0, 0) (1.2, 0.8) (2.8, 1.2) (4.4, 2.1) };
            
            \draw[orange, dashed, very thick] plot [smooth] coordinates { (0 + \hexshiftx, 0 - \hexshifty) (1.2 + \hexshiftx, 0.8 - \hexshifty) (2.8 + \hexshiftx, 1.2 - \hexshifty) (4.4 + \hexshiftx, 2.1 - \hexshifty) };
            
            \draw[blue, dashed, very thick] plot [smooth] coordinates { (0 - \hexshiftx, 0 + \hexshifty) (1.2 - \hexshiftx, 0.8 + \hexshifty) (2.8 - \hexshiftx, 1.2 + \hexshifty) (4.4 - \hexshiftx, 2.1 + \hexshifty) };

        \end{tikzpicture}
    }
    \caption{A hexagon, the angles of its sides, and shifted active corner-trajectories}
    \label{fig:hexagon-with-angles}
\end{figure}

Given an obstacle at point $(x_O, y_O)$, we can check if it is inside the unsafe region (or reachable set) in a computationally efficient fashion. 
If an obstacle lies outside the unsafe region, it would be either above both corner\hyp{}trajectories or below both corner\hyp{}trajectories for whichever corners are active. We can express the location of the obstacle with respect to a corner\hyp{}trajectory in a single equation by considering the value of $y_O - f(x_O - \Delta x_i) - \Delta y_i$ for active corner (vertex) $v_i$.
This term will be positive for both vertices $v_i, v_j$ if the obstacle is above both corner-trajectories, and similarly negative if the obstacle is below both. Therefore, any point $(x_O, y_O)$ in the safe region has a positive value for the product of the two expressions above, and any point in the unsafe region has a negative or zero value for this product.
This yields the following test to check if an object lies in (part of) the \textit{unsafe region}:
\begin{equation}
\left(y_O - f(x_O - \Delta x_i) - \Delta y_i\right) \cdot \left(y_O - f(x_O - \Delta x_j) - \Delta y_j\right) \leq 0
\label{eq:safe-region-test-corners-only}
\end{equation}
where $\Delta x_i, \Delta y_i, \Delta x_j, \Delta y_j$ are the (constant) offsets from the center of the polygon to the active corners (vertices) $v_i, v_j$.


When implementing this algorithm, the trajectories of all other vertices lie within the trajectories of the active corners, so to check whether an obstacle lies in the portion of the unsafe region defined by the active corners, it suffices to check over all pairs of vertices $(v_i, v_j)$ with $i, j \in \{1, 2, ... n\}$.
If the test in (\ref{eq:safe-region-test-corners-only}) indicates that an object lies within the unsafe region, trajectory $y=f(x)$ or $x=f(y)$ is clearly unsafe. 

\subsection{Notches at Transition Points Between Active Corners}
\label{sec:notches}

\tikzmath{\xstart = -4; \ystart = 3; 
\xnotch = 4; \ynotch=-3; 
\xend = 12; \yend = 3; 
\xex = 10; \yex = (\yend-\ynotch)/(\xend-\xnotch)^2 * (\xex - \xnotch)^2 + \ynotch;
\aleft = (\ystart-\ynotch)/(\xstart-\xnotch)^2;
\aright = (\yend-\ynotch)/(\xend-\xnotch)^2;
\ymin = \ynotch - \h - 0.5;
\ymax = \ystart + \h;}
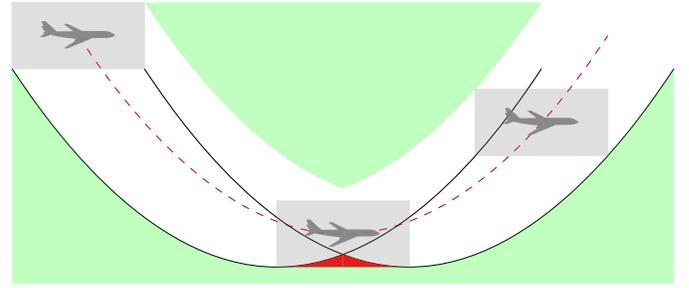
\begin{figure}
    \centering
    \begin{tikzpicture}[scale=0.44]
        \path[purple, dashed,draw] (\xstart, \ystart) parabola bend (\xnotch, \ynotch) (\xend, \yend) ;
        \path[black,draw] (\xstart + \w, \ystart - \h) parabola bend (\xnotch + \w, \ynotch - \h) (\xend + \w, \yend - \h);
        \path[black,draw] (\xstart - \w, \ystart - \h) parabola bend (\xnotch - \w, \ynotch - \h) (\xend - \w, \yend - \h);
        
        \fill [red, domain=\xnotch - \w:4, variable=\x]
          (\xnotch - \w, \ynotch - \h)
          -- plot ({\x}, {\aright*(\x - \xnotch + \w)*(\x - \xnotch + \w) - 3 - \h})
          -- (4, \ynotch - \h)
          -- cycle;
        \fill [red, domain=4:\xnotch + \w, variable=\x]
          (4, \ynotch - \h)
          -- plot ({\x}, {\aleft*(\x - \xnotch - \w)*(\x - \xnotch - \w) - 3 - \h})
          -- (\xnotch + \w, \ynotch - \h)
          -- cycle;

        \fill [green, nearly transparent, domain=\xstart - \w:\xnotch-\w, variable=\x]
          (\xstart - \w, \ymin)
          -- plot ({\x}, {\aleft*(\x - \xnotch + \w)*(\x - \xnotch + \w) - 3 - \h})
          -- (\xnotch-\w, \ymin)
          -- cycle;
          
        \fill [green, nearly transparent, domain=\xstart + \w:\xnotch, variable=\x]
          (\xstart + \w, \ymax)
          -- plot ({\x}, {\aleft*(\x - \xnotch - \w)*(\x - \xnotch - \w) - 3 + \h})
          -- (\xnotch, \ymax)
          -- cycle;

          
        \fill [green, nearly transparent, domain=\xnotch - \w:\xnotch + \w, variable=\x]
          (\xnotch - \w, \ymin)
          -- plot ({\x}, {\ynotch - \h})
          -- (\xnotch + \w, \ymin)
          -- cycle;
          
        \fill [green, nearly transparent, domain=\xnotch+\w:\xend + \w, variable=\x]
          (\xnotch+\w, \ymin)
          -- plot ({\x}, {\aright*(\x - \xnotch  - \w)*(\x - \xnotch - \w) - 3 - \h})
          -- (\xend + \w, \ymin)
          -- cycle;
          
        \fill [green, nearly transparent, domain=\xnotch:\xend - \w, variable=\x]
          (\xnotch, \ymax)
          -- plot ({\x}, {\aright*(\x - \xnotch  + \w)*(\x - \xnotch + \w) - 3 + \h})
          -- (\xend - \w, \ymax)
          -- cycle;

        \draw [fill, gray, nearly transparent] (\xstart -\w, \ystart -\h) rectangle (\xstart + \w, \ystart + \h); 
        \node [aircraft side,fill=gray,fill opacity=0.9,minimum width=1cm] at (\xstart + 0.2, \ystart) {};

        \draw [fill, gray, nearly transparent] (\xnotch -\w, \ynotch -\h) rectangle (\xnotch + \w, \ynotch + \h); 
        \node [aircraft side,fill=gray,fill opacity=0.9,minimum width=1cm] at (\xnotch + 0.2, \ynotch) {};
        
        \draw [fill, gray, nearly transparent] (\xex -\w, \yex -\h) rectangle (\xex + \w, \yex + \h); 
        \node [aircraft side,fill=gray,fill opacity=0.9,minimum width=1cm] at (\xex + 0.2, \yex) {};

    \end{tikzpicture}
    \caption{A rectangular airplane moving along a planar trajectory. At the transition point at the parabola's vertex, the ``notch" is visible and shaded in red; part of the object lies outside the corner-trajectories at this point.
    }
    \label{fig:tikz-plane-with-notch}
\end{figure}

It turns out that using only active corners would yield an underestimate of the reachable set, which would be unsound for verifying safety.
Figure \ref{fig:tikz-plane-with-notch} illustrates why: the white area bounded by the trajectories of the corners does not contain the red ``notch'', even though a collision would occur with an obstacle in this notch.
Therefore, if the test in (\ref{eq:safe-region-test-corners-only}) yields a value $> 0$, the advisory is not necessarily safe; we \textbf{additionally} check safety at all \textit{transition points} $(x_T, y_T)$ to see whether the obstacle at $(x_O, y_O)$ lies within the polygon centered at $(x_T, y_T)$. Recall that transition points are defined as points on the trajectory where the active corners switch.
In the full test for safety (Equation \ref{eq:safe-region-test-full}), this check is represented as \textbf{in\_polygon()} and can leverage one of many point-in-polygon implementations, which generally run in linear time on the order of number of vertices.
As the slope of the function may change at the boundary between piecewise subfunctions, we also add a notch check at each subdomain boundary.

\tikzmath{\sqshift=0.35;\sqend=3.0;\rectw=0.7;\recth=0.35;}
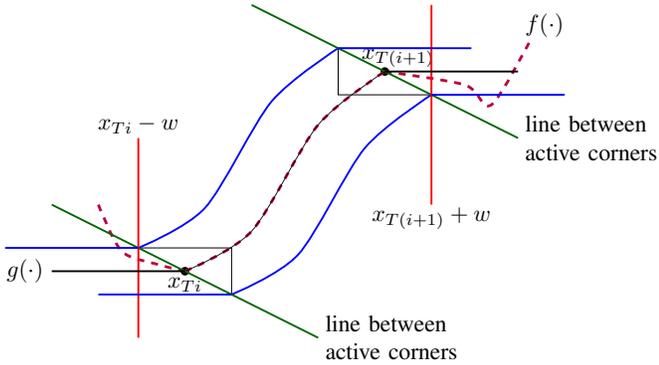
\begin{figure}
    \centering
    \resizebox{\columnwidth}{!}{%
        
        \begin{tikzpicture}
            \draw (-\rectw, -\recth) rectangle (\rectw, \recth);
            \draw (-\rectw + \sqend, -\recth + \sqend) rectangle (\rectw + \sqend, \recth + \sqend);
            
            \node[circle, fill, inner sep = 1.3pt] at (0,0) {};
            \node at (0, - 0.2) {$x_{Ti}$};
            
            \node[circle, fill, inner sep = 1.3pt] at (\sqend, \sqend) {};
            \node at (\sqend + 0.2, \sqend + 0.2) {$x_{T(i+1)}$};
    
            \draw[purple, dashed, very thick] plot [smooth] coordinates { (-1.3, 1.0) (-0.95, 0.3) (-0.4, 0.1) (0, 0)};
            \draw[purple, dashed, very thick] plot [smooth] coordinates { (0, 0) (1, 0.6) (2, 2.2) (\sqend, \sqend) };
            \draw[purple, dashed, very thick] plot [smooth] coordinates { (\sqend, \sqend) (3.4 + 0.8, 2.8) (3.8 + 0.8, 2.5) (4.4 + 0.8, 3.5) };
    
            \node at (4.4 + 0.8 + 0.2, 3.5 + 0.2) {$f(\cdot)$};
            
            \draw[opacity=0.9] plot [smooth] coordinates { (0, 0) (1, 0.6) (2, 2.2) (\sqend, \sqend) };

            \draw[blue, thick] plot [smooth] coordinates { (0.0 - \rectw, 0.0 + \recth) (1.0 - \rectw, 0.6 + \recth) (2.0 - \rectw, 2.2 + \recth) (\sqend - \rectw, \sqend + \recth) };
            
            \draw[blue, thick] plot [smooth] coordinates { (0.0 + \rectw, 0.0 - \recth) (1.0 + \rectw, 0.6 - \recth) (2.0 + \rectw, 2.2 - \recth) (\sqend + \rectw, \sqend - \recth) };
            
            \draw[black!60!green, thick] (-2.0, 1.0) -- (2.0, -1.0);
            \draw[black!60!green, thick] (-2.0 + \sqend, 1.0 + \sqend) -- (2.0 + \sqend, -1.0 + \sqend);
            \node[align=left] at (2.4 + 0.7, -1.0) {line between \\ active corners};
            \node[align=left] at (2.4 + 0.7 + \sqend, -1.0 + \sqend) {line between \\ active corners};

            \draw[red, thick] (-\rectw, 2) -- (-\rectw, -1);
            \draw[red, thick] (\sqend + \rectw, 4) -- (\sqend + \rectw, 1);
            \node at (-\rectw, 2 + 0.2) {$x_{Ti} - w$};
            \node at (\sqend + \rectw, 1 - 0.2) {$x_{T(i+1)} + w$};

            \draw[black, thick] (0.0, 0.0) -- (0.0 - 2.0, 0.0);
            \draw[black, thick] (\sqend, \sqend) -- (\sqend + 2.0, \sqend);
            \node at (-2.4, 0) {$g(\cdot)$};
    
            \draw[blue, thick] (0.0 - \rectw, 0.0 + \recth) -- (0.0 - 2.0 - \rectw, 0.0 + \recth);
            \draw[blue, thick] (0.0 + \rectw, 0.0 - \recth) -- (0.0 - 2.0 + \rectw, 0.0 - \recth);
            
            \draw[blue, thick] (\sqend - \rectw, \sqend + \recth) -- (\sqend + 2.0 - \rectw, \sqend + \recth);
            \draw[blue, thick] (\sqend + \rectw, \sqend - \recth) -- (\sqend + 2.0 + \rectw, \sqend - \recth);

        \end{tikzpicture}%
    }
    \caption{For piecewise functions, between transition points and/or piecewise boundaries, this figure shows the difference between $f(\cdot)$ and $g(\cdot)$ and the two additional checks on $(x_O, y_O)$ described in Section \ref{sec:piecewise}.}
    \label{fig:piecewise-traj-rules}
\end{figure}

\subsection{Handling Piecewise Functions}
\label{sec:piecewise}

In order to account for piecewise functions, we modify our method in two ways to avoid using a subfunction outside the subdomain over which it holds. The first is a modification to hold subfunctions constant outside of the subdomain over which they're defined; and the second is an additional boolean clause to the safety test in (\ref{eq:safe-region-test-corners-only}) so it only applies over a valid subdomain. In this case, the subdomain is an interval $[x_{Ti}, x_{T(i+1)}]$, where $x_{Ti}, x_{T(i+1)}$ may be piecewise boundaries \textit{or} transition points. Because of this, there may be many subdomains for a single piecewise case in which there happen to be many transition points.

First, we define a function $g(x)$ (or $g(y)$ symmetrically) that holds the value of each subfunction outside of its  subdomain $[x_{Ti}, x_{T(i+1)}]$. The function $g(\cdot)$ is used in place of $f(\cdot)$ in (\ref{eq:safe-region-test-corners-only}) above. Let $y_{Ti} = f(x_{Ti})$.
\begin{equation}
    g(x) = \begin{cases} 
        y_{Ti} & \text{if } x \leq x_{Ti} \\
        f(x) & \text{if }  x_{Ti} < x < x_{T(i+1)} \\
        y_{T(i+1)} & \text{if } x \geq x_{T(i+1)} \\
    \end{cases}
    \label{eq:def-g-x}
\end{equation}
Additionally, we add a clause to ensure the modified subfunction $g(\cdot)$ is only used over the correct subdomain. 
First, we check $x_{Ti} - w < x_O < x_{T(i+1)} + w$, where $w$ is the half-width of the object.
We also construct a line between the two active corners of the object in each of the two piecewise boundary locations $\big(x_{Ti}, y_{Ti}\big)$ and $\big(x_{T(i+1)}, y_{T(i+1)}\big)$ and check $(x_O, y_O)$ is between the two lines. This way, we ensure the test for being unsafe holds only for the region on which each subfunction applies. Figure \ref{fig:piecewise-traj-rules} illustrates the function $g(\cdot)$ and the additional subdomain-related clauses.


\tikzmath{\hexheight=1; \hexstartx=3.5; \curvex1=-2.0; \curvey1=0.3; \curvex2=1.5; \curvey2=-0.2;}
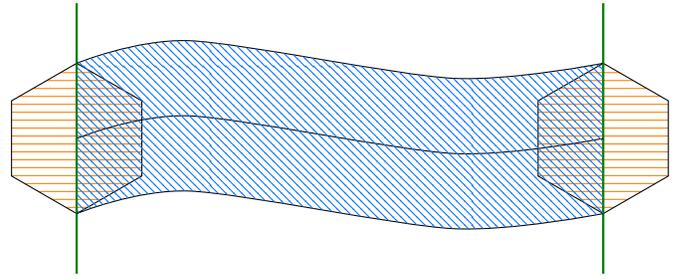
\begin{figure}
    \centering
    \begin{tikzpicture}[%
        hex/.style={draw, regular polygon, regular polygon sides=6,minimum size=2cm, shape border rotate=30}]
        
        \node[hex, pattern = {horizontal lines}, pattern color = orange] (pstart) at (-\hexstartx, 0) {};
        \draw[color=green!50!black, line width=0.3mm] (-\hexstartx, -1.8) -- (-\hexstartx, 1.8);
        \node[hex, pattern = {horizontal lines}, pattern color = orange] (pend) at (\hexstartx, 0) {};
        \draw[color=green!50!black, line width=0.3mm] (\hexstartx, -1.8) -- (\hexstartx, 1.8);

        \draw plot [smooth] coordinates { (-\hexstartx, 0)  (\curvex1, \curvey1) (\curvex2, \curvey2) (\hexstartx, 0) };
        \draw [name path=A] plot [smooth] coordinates { (-\hexstartx, 0 + \hexheight)  (\curvex1, \curvey1 + \hexheight) (\curvex2, \curvey2 + \hexheight) (\hexstartx, 0 + \hexheight) };
        \draw [name path=B] plot [smooth] coordinates { (-\hexstartx, 0 - \hexheight)  (\curvex1, \curvey1 - \hexheight) (\curvex2, \curvey2 - \hexheight) (\hexstartx, 0 - \hexheight) };
        \tikzfillbetween[of=A and B]{pattern = {north west lines}, pattern color = NavyBlue};
    \end{tikzpicture}
    \caption{Terms in \eqref{safe-region-test-full}, illustrated. \textcolor{green!50!black}{Green terms restrict test to relevant piecewise subdomains}, \textcolor{blue}{blue, diagonally-hatched terms check if obstacles are between active corner pairs}, and \textcolor{orange}{orange, horizontally-hatched terms check if obstacles are in the notch at transition points and subdomain bounds}.}
    \label{fig:full-equation-parts-colors}
\end{figure}

\begin{figure*}
\begin{equation}
    \label{eq:safe-region-test-full}
    \begin{split}
        \textbf{u}&\textbf{nsafe?} = \bigvee_{\{(g_i, x_{Ti}, x_{Tj})\}} \Bigg( \color{green!50!black} x_{Ti} - w < x_O < x_{Tj} + w \ \land \ \color{green!50!black} (x_O, y_O) \color{green!50!black} \text{ between } \\
        & \color{green!50!black}  \Big( \textbf{Line} \big( (x_{Ti} + \Delta x_i, y_{Ti} + \Delta x_i ), (x_{Ti} + \Delta x_j, y_{Ti} + \Delta x_j) \big), \color{green!50!black} \textbf{Line} \color{green!50!black} \big( (x_{Tj} + \Delta x_i, y_{Tj} + \Delta x_i ), (x_{Tj} + \Delta x_j, y_{Tj} + \Delta x_j) \big) \Big) \land  \\ 
        & \color{blue} \bigvee_{\{(v_i, v_j)\}}  \big(y_O - g(x_O +\Delta x_i) - \Delta y_i) \big(y_O - g(x_O +\Delta x_j) - \Delta y_j) \leq 0 \ \color{black} \Bigg) \lor \color{orange} \Big(\bigvee_{\{(x_{T}, y_{T})_i\}} \textbf{in\_polygon}(x_{Ti}, y_{Ti}, x_O, y_O) \Big)
    \end{split}
\end{equation}
\vspace{-13pt}
\end{figure*}

\subsection{Generic Explicit Formulation}

This leads to a generic quantifier-free explicit formulation to test whether an obstacle is in the safe region, where $\{(x_T, y_T)_i\}$ represents the set of all transition and boundary points on the trajectory between piecewise subdomains. Our algorithm yields a test for whether an obstacle is unsafe; negating the boolean formula or its result allows testing whether an obstacle lies in the safe region.

\eqref{safe-region-test-full} is color-coded in correspondence with Figure~\ref{fig:full-equation-parts-colors}. The first, third, and fourth lines ensure the test applies only over the correct piecewise domain and are in \textcolor{green!50!black}{green}; the second line checks the obstacle is between the active corners and is in \textcolor{blue}{blue}; the fifth line is in \textcolor{orange}{orange} and checks for the notch at transition points and piecewise subdomain boundaries.
For ease of presentation, we have focused on convex, centrally-symmetric objects and point-mass obstacles. However, our method extends to obstacles with area, to asymmetric, irregular polygons, and even non-convex objects, as detailed in Appendix~\ref{sec:extensions}.

\section{Proof of Equivalence}
\label{sec:proof}

\label{proof-section}
We prove the equivalence of the safe regions represented by 1) the implicit formulation and 2) our active-corner method for trajectories of form $y=f(x)$. The proof of soundness follows; the proof of completeness is in Appendix~\ref{sec:tightness-proof}, due to space constraints. The proof structure considers segments of the trajectory in which no active corner switch occurs; that is, where the angle of the tangent to the trajectory is bounded. In these segments, the bounds on the trajectory tangent angle allow us to bound the location of points in the interior of the polygon and show they lie between the two active corners. The two endpoints of a segment represent locations at which 1) the notch exists or 2) the trajectory switches to a new piecewise subfunction. In our method, these cases are handled by testing if obstacle $(x_O, y_O)$ is inside the polygon at various transition points $\big\{(x_T, y_T)\big\}_i$.

\subsection{Proof Preliminaries}
\label{proof-preliminaries}
Consider a segment of the motion along trajectory $y=f(x)$ or $x=f(y)$ in which no active corner switches or piecewise trajectory segment switches occur. We can arbitrarily rotate this segment of motion and the proof will hold, since the object translates along the trajectory without rotation. Assume, then, that a rotation is made by an angle $\theta$ such that the active corners are oriented along a vertical line. This rotation is an invertible transformation, so the logic of this proof holds through the entire trajectory. Because of this coordinate rotation, we consider only trajectories $y=f(x)$ for the proof; any trajectory $x=f(y)$ can be rotated into the form $y=f(x)$ invertibly, so our results hold for these forms as well. Let $v_i, v_j$ denote the active corners for this segment, with corresponding offsets $\Delta x_i, \Delta y_i, \Delta x_j, \Delta y_j$ in the rotated coordinate system.

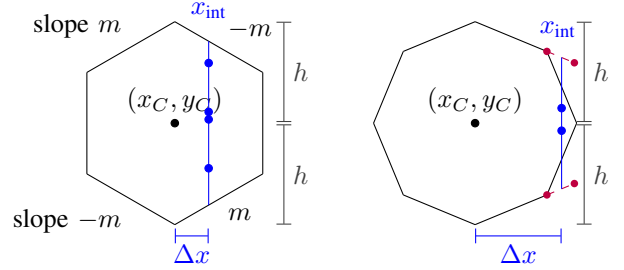
\begin{figure}
    \centering

   \begin{subfigure}[b]{0.45\columnwidth}

    \begin{tikzpicture}[scale=0.5]
      [thick]
      \newdimen\R
      \R=2.7cm
      \node [above] {$(x_C,y_C)$};
      \node [inner sep=1pt,circle,draw,fill] {};
      \foreach \x/\l in
        {30/,
         90/$-m$,
         150/slope $m$,
         210/,
         270/slope $-m$,
         330/$m$
        }
        \draw[postaction={decorate}] ({\x-60}:\R) -- node[auto,swap]{\l} (\x:\R);
    
        \newdimen\dX
        \dX=0.9cm
        \draw[blue] (\dX, -\R + 0.5773*\dX) -- (\dX, \R - 0.5773*\dX);
        \node[inner sep=1pt,blue, circle, draw, fill] at (\dX, 1.6cm) {}; 
        \node[inner sep=1pt,blue, circle, draw, fill] at (\dX, 0.3cm) {}; 
        \node[inner sep=1pt,blue, circle, draw, fill] at (\dX, 0.1cm) {}; 
        \node[inner sep=1pt,blue, circle, draw, fill] at (\dX, -1.2cm) {}; 
        
        \draw[darkgray, |-|] (\R + 0.2cm, -\R) -- (\R + 0.2cm, -0.03cm) node [midway, right] {$h$}; 
        \draw[darkgray, |-|] (\R + 0.2cm, 0.03cm)   -- (\R + 0.2cm, \R) node [midway, right] {$h$}; 
    
        \draw[blue, |-|] (0, -\R - 0.3cm) -- (\dX, -\R - 0.3cm) node [midway, below] {$\Delta x$};
        
        \node[blue] at (\dX, \R + 0.3cm) {$x_\text{int}$};
    \end{tikzpicture}
    \end{subfigure}
    \hfill
    \begin{subfigure}[b]{0.45\columnwidth}
    \begin{tikzpicture}[scale=0.5]
      [thick]
      \newdimen\R
      \R=2.7cm
      \node [above] {$(x_C,y_C)$};
      \node [inner sep=1pt,circle,draw,fill] {};
      \foreach \x/\l in
        {
            0/,
            45/,
            90/,
            135/,
            180/,
            225/,
            270/,
            315/            
        }
        \draw[postaction={decorate}] ({\x-45}:\R) -- node[auto,swap]{\l} (\x:\R);
        
        \draw[purple, dashed] (0.9239 * 0.76537*\R, \R - 0.38268 * 0.76537*\R) node[circle,fill,inner sep=1pt] {} -- ++(-22.5: 0.38268 * 0.76537*\R) node[circle,fill,inner sep=1pt] {};
        \draw[purple, dashed] (0.9239 * 0.76537*\R, -\R + 0.38268 * 0.76537*\R) node[circle,fill,inner sep=1pt] {} -- ++(22.5: 0.38268 * 0.76537*\R) node[circle,fill,inner sep=1pt] {} ;
        \newdimen\dX
        \dX=2.3cm
        \draw[blue] (\dX, -\R + 0.4142*\dX) -- (\dX, \R - 0.4142*\dX);
        \node[inner sep=1pt,blue, circle, draw, fill] at (\dX, 0.4cm) {}; 
        \node[inner sep=1pt,blue, circle, draw, fill] at (\dX, -0.2cm) {}; 
        
        \draw[darkgray, |-|] (\R + 0.2cm, -\R) -- (\R + 0.2cm, -0.03cm) node [midway, right] {$h$}; 
        \draw[darkgray, |-|] (\R + 0.2cm, 0.03cm)   -- (\R + 0.2cm, \R) node [midway, right] {$h$}; 
    
        \draw[blue, |-|] (0, -\R - 0.3cm) -- (\dX, -\R - 0.3cm) node [midway, below] {$\Delta x$};
        
        \node[blue] at (\dX - 0.1cm, \R - 0.2cm) {$x_\text{int}$};
    \end{tikzpicture}
    \end{subfigure}
    \caption{Figure of the slope of the sides of a regular hexagon and octagon.}
    \label{fig:hexagon-interior-dimensions}

\end{figure}

Since no active corner switch occurs, then we know the slope of function $y=f(x)$ is limited by the shape of the polygon itself -- let these bounds be $\pm m$, with $m$ representing the slope of the relevant sides of the polygon. Because the polygon is symmetric, the lower bound on slope is the negative of the upper bound (Figure~\ref{fig:hexagon-interior-dimensions}). \nishant{Add some explanation in the text here} This proof is presented considering regular, symmetric polygons for simplicity, but extends to asymmetric polygons as discussed in Appendix~\ref{sec:extensions}. To prove the soundness of our method, we must prove that all obstacles shown safe using our method ($\text{safe}_\text{expl}$) are also safe using the input implicit formulation ($\text{safe}_\text{impl}$). To prove $\text{safe}_\text{expl} \implies \text{safe}_\text{impl}$, we prove the contrapositive $\text{unsafe}_\text{impl} \implies \text{unsafe}_\text{expl}$.

Specifically, $\text{unsafe}_\text{impl}$ means that an obstacle at $(x_O, y_O)$ is inside a polygon centered at some coordinates $(x, y)$; $\text{unsafe}_\text{expl}$ means that the below holds from (\ref{eq:safe-region-test-corners-only}):
\begin{equation*}
\begin{split}
    \left(y_O - f(x_O - \Delta x_i) - \Delta y_i\right) \cdot \left(y_O - f(x_O - \Delta x_j) - \Delta y_j\right) \leq 0 \\ 
\end{split}
\end{equation*}

This proof has three sections: one holds for the majority of the trajectory segment, one for the beginning of the segment, and one of the end of the segment. The beginning- and end-of-segment proofs follow the form of the main proof but consider fixed polygons at the trajectory segment endpoints and are included in Appendix \ref{appendix-proofs} for space reasons.

\tikzmath{\hexheight=1; \hexstartx=4.5; \curvex1=-2.0; \curvey1=0.3; \curvex2=1.5; \curvey2=-0.2;}
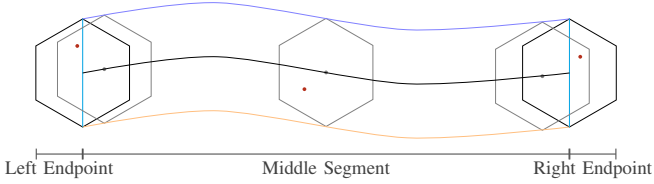
\begin{figure}
    \centering
    \resizebox{\columnwidth}{!}{%
    \begin{tikzpicture}[%
        point/.style={circle, radius=0.15cm, fill=black},
        hex/.style={draw, regular polygon, regular polygon sides=6,minimum size=2cm, shape border rotate=30}]
        
        \node[hex] (pstart) at (-\hexstartx, 0) {};
        \draw[color=Cerulean] (pstart.north) -- (pstart.south);

        \node[hex] (pend) at (\hexstartx, 0) {};
        \draw[color=Cerulean] (pend.north) -- (pend.south);

        \node[hex, color=gray] (pleft) at (-\hexstartx + 0.4, 0.07) {};
        \fill[color=gray] (pleft) circle (1pt) {};
        \fill[color=BrickRed] (-\hexstartx - 0.1, 0.5) circle (1pt) {};
        
        \node[hex, color=gray] (pmid) at (0, 0.01) {};
        \fill[color=gray] (pmid) circle (1pt) {};
        \fill[color=BrickRed] (-0.4, -0.3) circle (1pt) {};
        
        \node[hex, color=gray] (pright) at (\hexstartx - 0.5, -0.06) {};
        \fill[color=gray] (pright) circle (1pt) {};
        \fill[color=BrickRed] (\hexstartx + 0.2, 0.3) circle (1pt) {};
        
        \draw plot [smooth] coordinates { (-\hexstartx, 0)  (\curvex1, \curvey1) (\curvex2, \curvey2) (\hexstartx, 0) };
        \draw [color=blue!50] plot [smooth] coordinates { (-\hexstartx, 0 + \hexheight)  (\curvex1, \curvey1 + \hexheight) (\curvex2, \curvey2 + \hexheight) (\hexstartx, 0 + \hexheight) };
        \draw [color=orange!50] plot [smooth] coordinates { (-\hexstartx, 0 - \hexheight)  (\curvex1, \curvey1 - \hexheight) (\curvex2, \curvey2 - \hexheight) (\hexstartx, 0 - \hexheight) };
        
        \draw[|-|, color=darkgray] (-\hexstartx - 0.866, -1.5) -- node[midway, below] {Left Endpoint} (-\hexstartx, -1.5);
        \draw[|-|, color=darkgray] (-\hexstartx, -1.5) -- node[midway, below] {Middle Segment} (\hexstartx, -1.5);
        \draw[|-|, color=darkgray] (\hexstartx, -1.5) -- node[midway, below] {Right Endpoint} (\hexstartx + 0.866, -1.5);

    \end{tikzpicture}
    }
    \caption{Sections of proof}
    \label{fig:three-sections-proof}
\end{figure}

\subsection{Middle Segment Proof}
\label{middle-segment-proof}
Consider a (symmetric) polygon $P$, with half-height $h$ and half-width $w$, centered at $(x_C,y_C)$, where $y_C=f(x_C)$. Let $(x_\text{int}, y_\text{int})$ be a point inside or on the edges of $P$. We prove that interior point $(x_\text{int}, y_\text{int})$ lies between the active corners of an identical polygon $\overbar{P}$ located at $(x_\text{int}, f(x_\text{int}))$. We do this by bounding three terms: 1) $f(x_\text{int})$, 2) $y_\text{int}$, and 3) the active corners of $\overbar{P}$ to prove that $y_\text{int}$ lies between them. 


First, we bound $f(x_\text{int})$ (the center of $\overbar{P}$). Let $x_\text{int} = x_C + \Delta x$, for $\Delta x \in \left[-w, w \right]$. Let $f(x_\text{int})  = f(x_C + \Delta x) = y_C + \Delta y$, for some $\Delta y$ which we will bound. 
The slope of the trajectory $\frac{dy}{dx}$ is bounded by $(-m, m)$, because this proof considers a segment of motion with no changes in active corner. Hence $\Delta y$ is bounded proportionally to $\Delta x$, with $\Delta y \in \left ( - |m\Delta x|, |m\Delta x| \right)$. 
Therefore, $f(x_\text{int}) \in \left (y_C - |m\Delta x|, y_C + |m\Delta x| \right)$. Our proof proceeds assuming $\Delta x \neq 0$, since if $\Delta x = 0, x_\text{int}$ will lie on the vertical centerline of $P$. In that case, it is trivial to show $x_\text{int}$ lies between the active corners.

Now, we bound $y_\text{int}$. Recall $x_\text{int} = x_C + \Delta x$, for $\Delta x \in \left[-w, w \right]$.
Given that the slopes of the sides on the top and bottom of $P$ are $\pm m$, we assert that any $(x_\text{int}, y_\text{int})$ with $x_\text{int} = x_C + \Delta x$ has a corresponding $\Delta y_\text{int} \in \big[-h + |m\Delta x|, h-|m\Delta x|\big]$. This is illustrated in Figure \ref{fig:hexagon-interior-dimensions} with a hexagon, but it generalizes to any symmetric convex polygon. Given this, we can bound the $x$ and $y$ interior coordinates as below:




\begin{equation}
    \label{eq:interior-points-range}
    \left( x_\text{int}, y_\text{int} \right) = 
    \begin{bmatrix}
        x_C + \Delta x \\ 
        \left [y_C -h + |m\Delta x|, y_C + h - |m\Delta x| \right]
    \end{bmatrix}
\end{equation}

\begin{figure}
    \centering
    \resizebox{\columnwidth}{!}{%
        \begin{tikzpicture}[%
            hex/.style={draw, regular polygon, regular polygon sides=6,minimum size=2cm, shape border rotate=30}]
            
            \node[hex] (pstart) at (-\hexstartx, 0) {};
            \node[hex] (pend) at (\hexstartx, 0) {};
            \node[hex] (pmid) at (0, 0.01) {};
            \fill (pmid) circle (1pt) {};
            \node[hex, color=gray] (pshift) at (0.4, -.06) {};
            \fill[gray] (pshift) circle (1pt) {};
            \draw[color=ForestGreen] (pshift.north) -- (pshift.south);
            
            \node[] at ([yshift=0.25cm]pmid.north) {$x_C $};
            \node[] at ([yshift=-0.25cm]pshift.south) {$x_\text{int}$};
    
            \draw plot [smooth] coordinates { (-\hexstartx, 0)  (\curvex1, \curvey1) (\curvex2, \curvey2) (\hexstartx, 0) };
            \draw [color=blue!50] plot [smooth] coordinates { (-\hexstartx, 0 + \hexheight)  (\curvex1, \curvey1 + \hexheight) (\curvex2, \curvey2 + \hexheight) (\hexstartx, 0 + \hexheight) };
            \draw [color=orange!50] plot [smooth] coordinates { (-\hexstartx, 0 - \hexheight)  (\curvex1, \curvey1 - \hexheight) (\curvex2, \curvey2 - \hexheight) (\hexstartx, 0 - \hexheight) };
    
        \end{tikzpicture}
    }
    \caption{Shifted polygon illustration}
    \label{fig:shifted-polygon}
\end{figure}
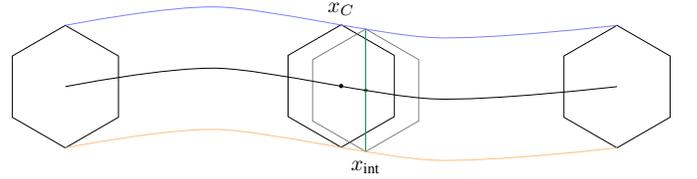

\eat{
Now we consider potential solutions to $f(x,y) = 0$ at an $x$-coordinate of $x_C+\Delta x$. Essentially, we attempt to bound $\Delta y$ in $f(x_C + \Delta x, y_C + \Delta y) = 0$. The slope of the trajectory \nishant{formalize trajectory/slope notion better?} is bounded by $(-m, m)$ because this proof considers a segment of motion with no changes in active corner. Therefore, $\Delta y_C$ for the new solution, or the movement in $y$, is bounded proportionally to $\Delta x$, with $\Delta y_C \in \big[ - |m\Delta x|, |m\Delta x| \big]$. 
}


Finally, we show interior point y-coordinate $y_\text{int}$ lies within the active corners of $\overbar{P}$. Because we consider a rotated coordinate frame such that the active corners are oriented along the vertical axis, the top and bottom active corners are located at $(\overbar{x_\text{top}}, \overbar{y_\text{top}}) = (x_\text{int}, f(x_\text{int})+h)$ and $(\overbar{x_\text{bot}}, \overbar{y_\text{bot}}) = (x_\text{int}, f(x_\text{int})-h)$, respectively. The bounds on $\overbar{y_\text{top}}$ and $\overbar{y_\text{bot}}$ are given by the following:
\begin{equation}
    \label{eq:top-bottom-corners-shifted}
    \begin{split}
        y_C - |m\Delta x| + h < \overbar{y_\text{top}} < y_C + |m\Delta x| + h \\ 
        y_C - |m\Delta x| - h < \overbar{y_\text{bot}} < y_C + |m\Delta x| - h
    \end{split}
\end{equation}

Then $y_\text{int} \leq y_C - |m\Delta x| + h < \overbar{y_\text{top}}$ and $y_\text{int} \geq y_C + |m\Delta x| - h > \overbar{y_\text{bot}}$.

The top active corner trajectory is given by
    $f_\text{top}(x) = f(x) + h$
and the bottom active corner trajectory is given by
    $f_\text{bot}(x) = f(x) - h$.
By definition,
        $f_\text{top}(x_\text{int}) = \overbar{y_\text{top}}$ and 
        $f_\text{bot}(x_\text{int}) = \overbar{y_\text{bot}}$, 
or equivalently,
        $\overbar{y_\text{top}} - f_\text{top}(x_\text{int}) = 0$ and
        $\overbar{y_\text{bot}} - f_\text{bot}(x_\text{int})= 0$.
Since $y_\text{int} < \overbar{y_\text{top}}$ and $y_\text{int} > \overbar{y_\text{bot}}$, 

\begin{equation}
        y_\text{int} - f_\text{top}(x_\text{int}) < 0 \qquad\qquad
        y_\text{int} - f_\text{bot}(x_\text{int}) > 0
    \label{eq:top-bottom-interior-fn}
\end{equation}

By multiplying the equations in (\ref{eq:top-bottom-interior-fn}), we get
\begin{equation}
    \label{eq:explicit-at-interior-point}
    (y_\text{int} - f_\text{top}(x_\text{int})) \cdot (y_\text{int} - f_\text{bot}(x_\text{int})) < 0
\end{equation}

This is equivalent test for whether an object lies in the unsafe region from (\ref{eq:safe-region-test-corners-only}). Therefore we have shown that for all $(x_C,y_C)$ points satisfying $y_C = f(x_C)$, all points $(x_\text{int}, y_\text{int})$ inside and on the boundary of a polygon centered at $(x_C,y_C)$ also have $(f_\text{top}(x_\text{int}) - y_\text{int}) \cdot (f_\text{bot}(x_\text{int}) - y_\text{int}) \leq 0$. These are exactly the definitions of $\text{unsafe}_\text{impl}$ and $\text{unsafe}_\text{expl}$ from \ref{proof-preliminaries} earlier. 
Therefore, we have shown $\text{unsafe}_\text{impl} \implies \text{unsafe}_\text{expl}$ and the contrapositive $\text{safe}_\text{expl} \implies \text{safe}_\text{impl}$ holds as well.

        

        




\section{Implementation}
\label{sec:implementation}

We have implemented our automated method in Python using SymPy, a symbolic math library \cite{SymPyPaper}.
The code implementing our algorithm and the applications in Section \ref{sec:applications} is available on 
GitHub at \url{https://github.com/nskh/automatic-safety-proofs}.

Given a fully symbolic trajectory and object, we first identify the angles corresponding to sides of the object. Then, following Section \ref{sec:active-corners}, we identify points on the trajectory corresponding to the angles $\theta_i$ of each side of the object. To avoid discontinuities in the $\arctan$ function, the implementation solves a reformulation: $\frac{\partial f}{\partial x} \sin(\theta_i) = \frac{\partial f}{\partial y} \cos(\theta_i)$. 
Solving this equation may yield either $y$ in terms of $x$ or the reverse. 
In this case, we substitute the implicit solution for $x$ or $y$ into the trajectory equation, eliminate the remaining variable, and identify transition points.
Given transition points, we can implement the test from (\ref{eq:safe-region-test-full}). Sympy includes a ``point-in-polygon" method, which we use to identify if an obstacle $(x_O, y_O)$ lies in the ``notch" at any transition point. The output explicit formulation can be expressed either in \LaTeX{} or in Mathematica format; output in Mathematica also supports generating code to copy-paste directly and plot the safe region using Mathematica's \texttt{RegionPlot[]} functionality. Examples can be found in Figures \ref{fig:tacas-acas-paper-figure} and \ref{fig:eytan-paper-figure}.

In order to implement our method in a fully symbolic fashion, we must account for the potential values of symbols when instantiated. We can leverage Sympy's built-in ``assumptions" to specify that certain symbols representing, say, trajectory parameters or object dimensions are real, positive, and/or nonzero, but these assumptions may not suffice to construct a fully symbolic safe region. In that case, our fully symbolic implementation computes a number of potential valid safe regions. As detailed in Section \ref{sec:piecewise}, we construct the explicit formulation using many clauses defined on intervals between transition points and/or piecewise boundaries. In the symbolic case, the order of these terms may differ, depending on, say, the sign of a variable in the trajectory. Additionally, symbolic piecewise cases for, say, $x < b$ may mean that certain transition points do not occur at all if $b$ lies in some range. Correspondingly, our fully symbolic implementation computes all valid orderings of piecewise boundaries and transition points; it additionally considers all valid combinations of transition points to account for ``notches" that may not exist when piecewise bounds and/or trajectory parameters are instantiated. In order to check if orderings are valid, we attempt to sort using the Sympy assumptions: if we know $b$ is positive, no returned ordering will place $b$ before a transition point at $0$, for example. Additionally, we enforce that adjacent points in the ordering ``come from" the same functions: we will not return an ordering where a transition point from piecewise subfunction $f_1$ lies between the piecewise boundaries for $f_2$. Doing so ensures that we generate relatively few ($\sim$ 10) potential orderings despite considering many combinations, though examples with intractably many orderings do exist.

\section{Applications and Evaluation}
\label{sec:applications}


\subsection{Verification of vertical maneuvers in ACAS~X}
\label{acas-application}

\begin{figure}
    \centering
    \includegraphics[width=\columnwidth]{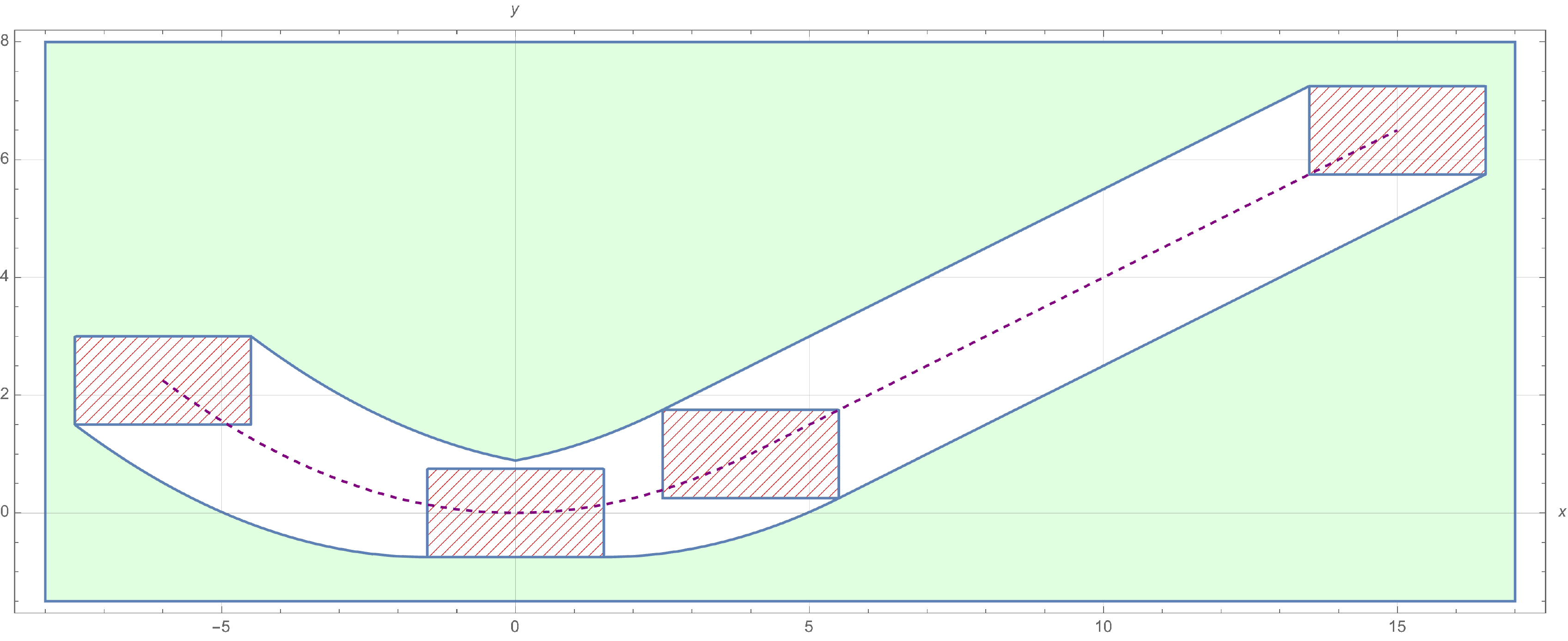}
    \caption{Safe region for an instance of \cite{jeannin2015acastacas}. The notches are the red-hatched rectangles and the trajectory is dashed in purple.}
    \label{fig:tacas-acas-paper-figure}
\end{figure}

A collision avoidance system intended to prevent near mid-air collisions, ACAS~X, was verified in \cite{jeannin2015acastacas}. The KeYmaera~X proof presented in \cite{jeannin2015acastacas} required a significant amount of human interaction (on the order of hundreds of hours), while the method presented in this paper generates an explicit formulation from the trajectory fully automatically. ACAS~X prevents collisions between aircraft by issuing advisories (control commands) to one aircraft, the \textit{ownship}. The bounds of aircraft in this work are shaped like hockey pucks (cylinders wider than they are tall) of a radius $r_p$ and half-height $h_p$. From a side perspective of an encounter between aircraft, the bounds are rectangular. In \cite{jeannin2015acastacas}, verification was performed in a side-view perspective, assuming two aircraft approach each other in a vertically-oriented planar slice of three dimensions. A careful choice of reference frame can reduce a three-dimensional encounter between aircraft into a two-dimensional system, by modeling the encounter as a 1-dimensional vertical encounter and the distance of a horizontal encounter \cite[Section~6]{jeannin2015acastacas}.

To simplify calculations, \cite{jeannin2015acastacas} used the relative horizontal speed $r_v$ of the two aircraft and assumed it constant; the vertical velocity of the oncoming aircraft $\dot{h}$ is also assumed constant. Advisories consist of climb and descent speed advisories, yielding ownship trajectories that are piecewise combinations of parabolas and straight lines. One example trajectory is below in (\ref{eq:acas-trajectory}), which assumes the advisory issued is for the ownship to climb at a rate $\dot{h}_f$ greater than its current vertical velocity $\dot{h}_0$. $(r_t, h_t)$ are the $(x,y)$ coordinates for trajectory $\TT$ in this example, and $a_r$ is the acceleration.

\begin{equation}
    \label{eq:acas-trajectory}
    \left( r_t, h_t \right) = \begin{cases} \left(r_v t, \frac{a_r}{2}t^2 + \dot{h}_0 t \right) & \text{ for } 0 \leq t < \frac{\dot{h}_f - \dot{h}_0}{a_r} \\ 
    \left(r_v t, \dot{h}_f t - \frac{(\dot{h}_f - \dot{h}_0)^2}{2a_r} \right) & \text{ for } \frac{\dot{h}_f - \dot{h}_0}{a_r} \leq t
    \end{cases}
\end{equation}
The implicit formulation of the safe region is below, for an oncoming aircraft at relative coordinates $r_O, h_O$.

\begin{equation}
    \label{eq:acas-tacas-paper-implicit}
    \forall t. \forall r_t. \forall h_t. \big( (r_t, h_t) \in \TT \implies |r_O-r_t| > r_p \vee |h_O - h_t| > h_p \big)
\end{equation}

In \cite{jeannin2015acastacas}, the authors eliminate the parametrization over $t$, which yields an initial parabolic section and then straight-line motion after. We use this $t$-free trajectory to compute the unsafe region, which is displayed in Figure \ref{fig:tacas-acas-paper-figure}. A boolean formulation of the unsafe region is in Appendix \ref{appendix-tacas}.

\subsection{Verified Turning Maneuvers for Unmanned Aerial Vehicles}
\label{sec:uav-application}

\begin{figure}
    \centering
    \includegraphics[width=0.8\columnwidth]{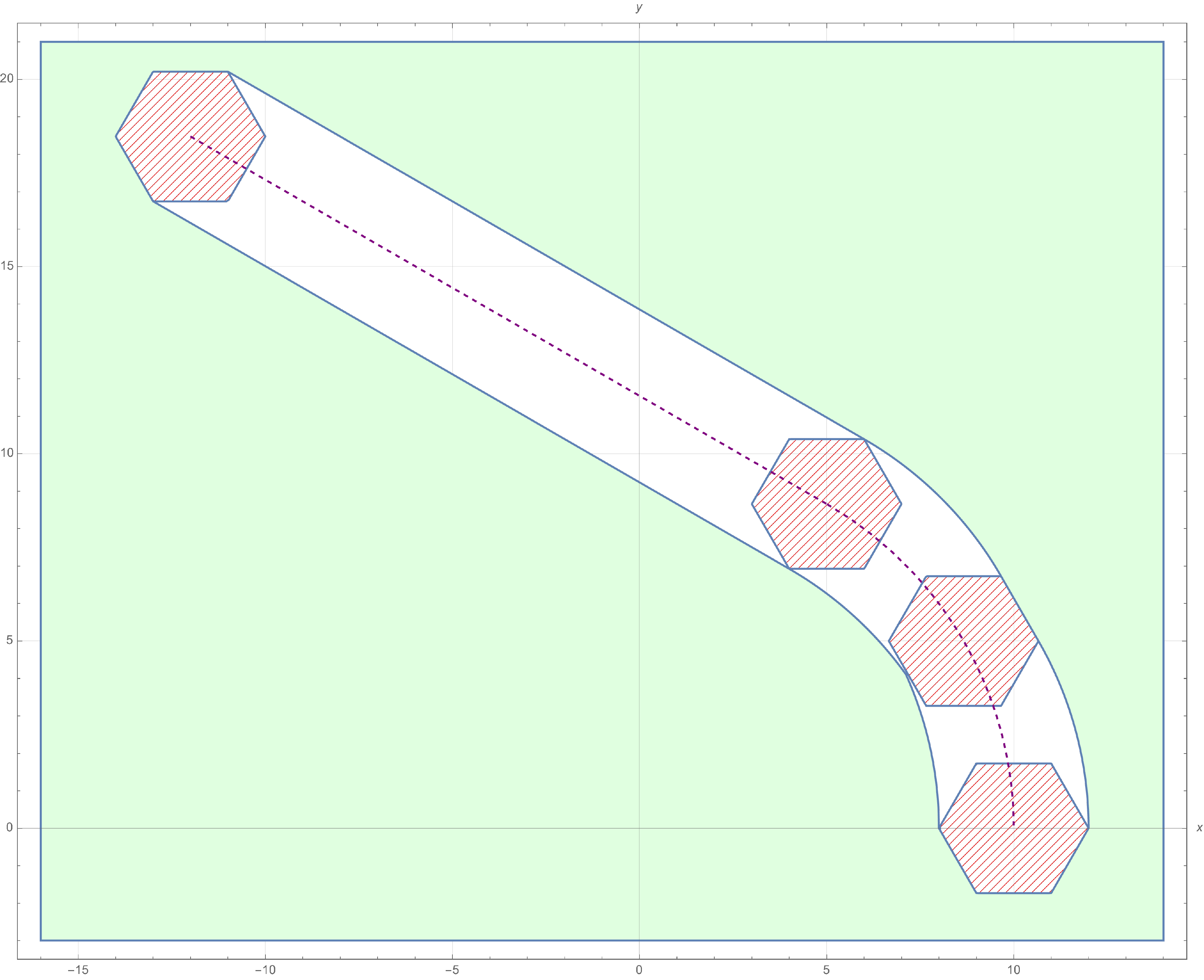}
    \caption{Approximated safe region for an instance of \cite{adler2019formal}. The notches are the red-hatched hexagons, the trajectory is dashed in purple.}
    \vspace{-4pt}
    \label{fig:eytan-paper-figure}
\end{figure}

Turning maneuvers for unmanned aerial vehicles (UAVs) have been verified as safe in \cite{adler2019formal}, where the UAV was represented as a circular safety buffer around a point object fixed along the trajectory. The KeYmaera X proof presented in \cite{adler2019formal} required a significant amount of human interaction (on the order of hundreds of hours); in contrast, the method presented in this paper generates an explicit formulation from the trajectory fully automatically. This work represents motion in a two-dimension plane viewed top-down, with the buffer ``puck" taking the form of a circle. The turning maneuver trajectory moves along a circular arc then in a straight line:

\begin{equation}
    \label{eq:eytan-paper-trajectory}
    (x_\TT, y_\TT) = 
    \begin{cases}
    x_\TT^2 + y_\TT^2 = R^2 & y_\TT < x_\TT \tan\theta \\
    y_\TT = \frac{R\cos\theta - x_\TT}{\tan\theta} + R\sin\theta & y_\TT \geq x_\TT \tan\theta \\
    \end{cases}
\end{equation}

With a circular safety buffer of radius $r_p$, the implicit formulation ensures for all points along the trajectory, the obstacle $(x_O, y_O)$ is at least $r_p$ away.
\begin{equation}
    \label{eq:eytan-paper-implicit}
    \forall x_\TT. \forall y_\TT. \big( \text{traj}(x_\TT, y_\TT) \implies (x_O - x_\TT)^2 + (y_O - y_\TT)^2 \geq r_p^2 \big)
\end{equation}

Note that our method does not support circular objects, only polygons, so we overapproximate the circular safety buffer as a regular hexagon inscribing a circle. This approximation allows a valid overapproximation of the unsafe region, since the hexagon contains the original circle in \cite{adler2019formal}. Note that the approximation of a circle can be made arbitrarily precise by increasing the number of sides of a polygon used. A plot of the unsafe region is in Figure \ref{fig:eytan-paper-figure} and a boolean formulation of the unsafe region is in Appendix \ref{appendix-uav}.

\begin{table*}
\centering
\begin{tabular}{|c|c|c|c|c|c|}
    \hline \textbf{Example} & \textbf{Instance} & \textbf{Active Corners Time} & \textbf{Active Corners RAM} & \textbf{CAD Time} & \textbf{CAD RAM}\\ \hline
    UAV     & Fully Numeric         & \textbf{0.48 sec } & \textbf{7.1 MB} & \emph{2381* sec} & 30.89 MB \\ 
    UAV     & Numeric Trajectory    & \textbf{0.82 sec } & \textbf{8.4 MB} & DNF                & 50+ GB \\ 
    UAV     & Numeric Hexagon       & \textbf{38 sec   } & \textbf{22 MB } & DNF                & 100+ GB \\ 
    UAV     & Fully Symbolic        & \textbf{45 sec   } & \textbf{24 MB } & DNF                & 100+ GB \\ \hline
    Dubins  & Fully Numeric         & \textbf{1.2 sec  } & \textbf{9.0 MB} & DNF                & 11+ GB \\
    Dubins  & Fully Symbolic: Rectangle  & \textbf{4505 sec}  & \textbf{91 MB} & DNF            & 4+ GB \\
    Dubins  & Fully Symbolic: Hexagon    & DNF  & N/A    & DNF                & 8+ GB \\ \hline
    ACAS X  & Fully Numeric         & 0.13 sec  & 5.9 MB & \textbf{0.04 sec}           & \textbf{160 KB} \\ 
    ACAS X  & Numeric Trajectory    & 0.48 sec  & 6.6 MB & \textbf{0.04 sec}           & \textbf{188 KB}  \\ 
    ACAS X  & Numeric Rectangle     & 0.51 sec  & 6.6 MB & \textbf{0.2 sec }           & \textbf{325 KB} \\ 
    ACAS X  & Fully Symbolic        & \textbf{0.57 sec}  & 6.6 MB & 1.1 sec            & \textbf{1.8 MB} \\ \hline
\end{tabular}
\caption{Evaluation results, with better results \textbf{bolded}. DNF: example did not finish in 8+ hours. \emph{*: incorrect answer.}} 
\vspace{-2pt}
\label{evaluation-table}
\end{table*}

\subsection{Runtime Evaluation}
\label{runtime-eval}
This section presents a comparison of our method to quantifier elimination via cylindrical algebraic decomposition (CAD) in a variety of cases, from fully numeric to fully symbolic \cite{CADQEPaperCollins}. Results were generated using a 2017 iMac Pro workstation with 128\,   GB of RAM, with CAD results using Mathematica's \texttt{Resolve} implementation. A table of results is shown in Table \ref{evaluation-table}. We use the examples from \ref{acas-application} (ACAS X) and \ref{sec:uav-application} (UAV). The Dubins path example is inspired by common path planners     and takes the form of two circular arcs connected by a straight line and ending with a line; its symbolic trajectory equation is included in Appendix \ref{appendix-dubins}.

Our findings in Table \ref{evaluation-table} demonstrate the advantages and disadvantages of our method relative to quantifier elimination using CAD. For non-polynomial examples like a rectangle moving along the Dubins path described above or the UAV example from \cite{adler2019formal}, the active corner method is able to compute fully symbolic formulations of the safe region when CAD fails to return an answer when run overnight (8+ hours). We do note that due to the complexity of a symbolic \textit{hexagon} moving along the Dubins path, the number of transition points means our method cannot compute an answer, though neither can CAD. For a fully numeric example from \cite{adler2019formal}, CAD took 2381 seconds to run but returned \texttt{False} incorrectly in place of a region. Additionally, memory is often a constraint for symbolic computation given the CAD algorithm's doubly-exponential runtime \cite{QEDoublyExpDavenport}; many examples consumed 100+ GB of RAM and one case grew to consume 350 GB of RAM without returning an answer. In the worst case, however, our method consumes under 100MB of RAM. On the other hand, for strictly polynomial examples like that in \cite{jeannin2015acastacas}, CAD runs quickly and efficiently, though our method remains competitive. 

\eat{
Notes from papers: 
\begin{itemize}
    \item give lines of code, status
    \item input
    \item process of accepting input and running code
    \item benefits with respect to efficiency, performance
\end{itemize}
SMT Solvers
\begin{itemize}
    \item give rules and examples
    \item address possible classes of input
    \item talk about advantages of doing it this way
    \item talk about potential problems and how they are addressed
\end{itemize}
Path Feasibility
\begin{itemize}
    \item give overview of what all the section has
    \item helper functions and definitions
    \item show code and then explain it
    \item if written out, can write it as 2 cases or phases, can have steps with numbers
\end{itemize}
Frenetic
\begin{itemize}
    \item sentence or two talking about how not only have we found our algorithm, we have also implemented it
    \item give language, level of progress, organization/location
    \item for each key part, talk about what its job is
    \item mention key data structures as applicable
    \item give algorithm (pseudo-code or written out or both?)
    \item talk about what the code can handle
    \item talk about future extensions on the code
\end{itemize}
}

\section{Related Work}
\label{sec:related-work}

Reachability computation is a vital question in safety-critical cases where users seek to guarantee properties or behavior. 
One method of constructing reachable sets for safety is zonotope reachablity \cite{althoff2021setpropagation-survey}. Reachability computation using zonotopes offers efficient algorithmic methods and supports analysis of dynamical systems with uncertainty. Zonotopes have been used in verification of automated vehicles \cite{althoff2014online}, the design of safe trajectories for quadrotor aircraft \cite{kousik2019safe}, and the analysis of power systems \cite{el2017compositional}, among other applications. 
Zonotope reachability methods discretize a dynamical system and iteratively propagate an estimate of the reachable set forward in time. Their input is a differential equation, while our method requires an explicit closed-form trajectory. For the purpose of checking safety, the estimate of the reachable set must be either exact or an overestimate; in order to deal with discretization error, zonotope methods repeatedly overestimate the reachable interval. Zonotope methods for nonlinear systems rely on linearization and again account for error that may occur by expanding the reachable set \cite{ZonotopesNonlinearAlthoff}. Our method yields exact reachable sets. 
While it is possible to model convex object reachability with zonotopes, the reachable set expands with the time horizon because the dimensions of the object are treated not as constant dimensions but as uncertainty in initial conditions that is propagated forward through time \cite{girard2005reachability}. 
Set-valued constraint solving may be used but similarly relies on inexact discretization \cite{jaulin2012solving}.
Other reachability methods for differential equations include Hamilton-Jacobi reachability for systems with complex, nonlinear, high\hyp{}dimensional dynamics \cite{bansal2017hamilton}, and control barrier functions, which enable the construction of safe optimization\hyp{}based controllers \cite{ames2019control}. 

A counterpart to reachability is automatic invariant generation for hybrid systems, in which a formal statement showing a system never evolves into an unsafe state is proved.
In \cite{ghorbal2014characterizing}, the authors proved a polynomial and its Lie derivatives can represent algebraic sets of polynomial vector fields. A procedure to check invariance of polynomial equalities was proposed in \cite{ghorbal2014invariance}.
Semi-algebraic invariants for polynomial ODEs were studied in \cite{ghorbal2017hierarchy,sogokon2016method,liu2011computing}.
Invariants for hybrid systems were studied in \cite{sankaranarayanan2004constructing} and \cite{matringe2010generating}. Relational abstractions bridge the gap between continuous and discrete modes by over-approximating continuous system evolution to summarize the system as a purely discrete one using invariant generation \cite{sankaranarayanan2011relational}. Barrier certificates have also been used as invariants for safety verification in hybrid systems \cite{prajna2004safety}. 

Our work has similar aims to swept-volume collision checking, from path planning and graphics, in which approximate, efficient collision-checking is performed as a volume is moved along a path. A convex over-approximation swept-volume approach was presented in \cite{foisy1993safe}. Swept-volume checking in four dimensions was performed using an intersection test in space-time in \cite{cameron1990collision}. An efficient algorithm computing distances between convex polytopes, the Lin-Canny algorithm, was proposed for this task in \cite{lin1992efficient,lin1993efficient}. 
Methods are typically discrete and approximate for performance in online applications. 
That said, there are some exact methods such as collision checking for straight-line segments like those on robotic arms \cite{schwarzer2004exact} and an algorithm for large-scale environments \cite{cohen1995collide}. However, these methods operate on individual collision checking instances, such as graphics simulations or video game environments, and their results cannot be used repeatedly. Our method yields provably correct, fully symbolic, and exact safe regions for continuous trajectories and supports, for example, quantifier-free and efficient testing in runtime or in large-scale settings once a desired safe region formulation has been generated.

Another alternative to this work is quantifier elimination, \cite{tarski1951decision,CADQEPaperCollins} a general algorithm for converting formulas with quantified variables into equivalent statements that are quantifier-free. Quantifier elimination can be performed using Cylindrical Algebraic Decomposition (CAD), an algorithm that operates on semialgebraic sets \cite{CADQEPaperCollins,arnon1984cylindrical}. QEPCAD is one notable software tool implementing CAD that could be used in this work \cite{QEPCADBrown2003}. The runtime of the CAD algorithm is doubly exponential in the number of total variables (not the number of quantified variables) \cite{QEDoublyExpDavenport,brown2007complexity}; we offer a detailed comparison to CAD in Mathematica in Section \ref{runtime-eval}.

\section{Conclusion and Future Work}
\label{sec:discussion}

%
We would like to study how our method extends to objects translating in 3 or $n$ dimensions; trajectories in the form of inequalities; rotating objects; and invariants of differential equations of the form $f(x,y)=0$ rather than explicit trajectories. On the implementation side, we envision an extension to automatically output a machine-checkable proof of equivalence between the implicit and explicit formulations, in a theorem prover such as Coq, PVS, Isabelle, or KeYmaera~X.

\bibliographystyle{plain}
\bibliography{mybib}


\appendices



\newpage
        
\section{Safe Region Inequalities For Figure \ref{fig:safe-region-example}}
\label{fig-1-inequalities}
These inequalities are plotted in Figure \ref{fig:safe-region-example}. The \textit{safe region} is simply the negation of \eqref{logic-explicit-safe-example}.
\begin{equation}
    \label{eq:logic-explicit-safe-example}
    \begin{split}
        \Big(& (x_O \geq -w) \  \land \ (y_O \leq h)   \ \land  (y_O \geq -2x_O -2w - h) \\ &\land \ (x_O \leq 5+w) \ 
         \land \ (y_O \leq -2x_O + 2w + h)   \\ &\land  \ (y_O \geq -10-h)\Big) \ \bigvee
        \Big((x_O \geq 5-w) \\ 
        &\land \ (y_O \geq -10-h) \ \land  (y_O \leq x_O + w + h - 15)   \\ & \land \ (y_O \geq x_O - w - h - 15)\Big)
    \end{split}
\end{equation}
Note the first disjunction in \eqref{logic-explicit-safe-example} corresponds to the motion of the aircraft on the left side of Figure~\ref{fig:safe-region-example} as it descends; the second corresponds to the right side as the aircraft ascends.

\section{Extensions}
\label{sec:extensions}

In the main body of the paper we have considered point-mass obstacles, but the reasoning extends to obstacles that have the same properties as the object (convex and centrally symmetric).
This is achieved through a reduction where the shape of the obstacle is incorporated into the shape of the object.
For example, in the simple case where both are horizontal rectangles, with the object of height $2h$ and width $2w$, and the obstacle of height $2h_O$ and width $2w_O$, the object and obstacle intersect if and only if the center of the obstacle is contained in a virtual object with the same center as the initial object, but of height $2(h+h_O)$ and width $2(w+w_O)$. We have thus reduced the problem of collision avoidance with a convex object to a problem avoidance with a point-mass object.
A similar reasoning -- albeit a little more complicated -- can be applied to any convex, centrally symmetric obstacle.

For ease of of presentation, and because they appear in most practical applications, we have focused on objects that are convex and centrally symmetric. We can extend the reasoning to non-centrally symmetric objects: the only difference in that case is that pairs of active corners do not change together, but rather one active corner may change on one side, and another active corner may change on the other side later. Pairs of active corners are thus not opposite corners of the object anymore. The convexity of the object (and obstacles) is essential for active corners; however, we can extend our reasoning to non-convex, polygonal objects by seeing them as unions of convex sub-objects and ensuring collision avoidance with each sub-object. Finally, due to its reliance on corners, our method cannot handle circles or ellipses, but they can be approximated by polygons.

\section{Proofs}
\label{appendix-proofs}
\subsection{Left Endpoint Proof}
\label{initial-segment-proof}
Assume the coordinate rotation is paired with a shift such that the trajectory segment goes from $(0,0)$ to $(x_F, y_F)$. The section above showed interior points with $x_\text{int} \in [0, x_F]$ lie between the active corner-trajectories, but polygons near the endpoints may have interior points with $x_\text{int} < 0$ or $x_\text{int} > x_F$, despite being centered along the trajectory segment. Here, we consider interior points with $x$-coordinates that are less than 0 and show these points lie within the \textit{initial polygon} centered at $(0,0)$. For these interior points, polygon center $x_C \leq w$. 

First, bound the $y$-coordinate of such polygons: with minimum trajectory slope $-m$ and maximum $+m$, $y_C \in (-mx_C, mx_C)$. Now we can bound the locations of the interior points of these polygons, using a result from the prior section. For an interior point with $x$-axis offset $\Delta x$, we have corresponding y-axis offset $\Delta y \in [-h + |m \Delta x|, h - |m\Delta x|]$. Therefore, interior points of the polygon at $x_C +\Delta x$ have $y_\text{int} \in (-m x_C -h + |m \Delta x|, m x_C + h - |m\Delta x|)$. Note that because $x_\text{int} < 0, \Delta x < -x_C$. If $x_\text{int} > 0$, the proof in Section \ref{middle-segment-proof} applies. $x_\text{int} = 0$ is a trivial case since all interior points there are inside the initial polygon centered at $(0,0)$.

Now we show that $(x_\text{int}, y_\text{int})$ always lies in the polygon centered at $(0, 0)$. First we compare $x_\text{int}$ to the polygon center $0$ to determine the $x$-axis offset $\Delta x_\text{init}$, relative to the initial polygon at $(0,0)$. We have offset 
$|x_\text{init}| = |\Delta x| - x_C$. By the same logic used to bound $y_\text{int}$ in Section \ref{middle-segment-proof}, $\Delta y_\text{init} \in [-h + m * (|\Delta x| - x_C), h - m * (|\Delta x| - x_C)]$. 
Expanding this yields $\Delta y_\text{init} \in [-h + |m \Delta x| - m x_C, h - m |\Delta x| - m x_C]$.  If we compare $y_\text{int}$ and $y_\text{init}$, we see $y_\text{int}$ and $y_\text{init}$ fall within the same ranges (the former is open and the latter is closed). Therefore, any interior point $(x_\text{int}, y_\text{int})$ of a polygon with $x_\text{int} < 0$ lies inside the initial polygon centered at $(0,0)$.


\subsection{Right Endpoint Proof}
This proof proceeds similarly to the proof in Section \ref{initial-segment-proof}, though in the neighborhood of endpoint $(x_F, y_F)$ (at which a piecewise change or active corner switch may occur). We show that any interior point of a polygon on trajectory $y=f(x)$ with an $x$-coordinate $x_\text{int} > x_F$ lies within the polygon centered at $(x_F, y_F)$.

Consider a polygon centered at some $(x_C, y_C)$ that has some interior points beyond $x_F$: its distance in the $x$-axis from $x_F$ is $x_F - x_C$ and therefore $y_C \in (y_F - m (x_F - x_C), y_F + m (x_F - x_C))$. Now, as in \ref{initial-segment-proof}, we consider interior points of this polygon at some positive $\Delta x$ offset from the center. Using similar logic as before, we find $\Delta y \in [-h + m\Delta x, h - m\Delta x]$. For $x_\text{int} \triangleq x_C + \Delta x$, we have $y_\text{int} \in (y_F - m (x_F - x_C) - h + m\Delta x, y_F + m (x_F - x_C) +  h - m\Delta x)$.

Now we show that $(x_\text{int}, y_\text{int})$ always lies in the polygon centered at $(x_F, y_F)$. First we must compare $x_\text{int}$ to $x_F$ to determine the $x$-axis offset $\Delta x_\text{final}$, relative to the final polygon at $(x_F, y_F)$. This offset $\Delta x_\text{final} = \Delta x - (x_F - x_C) = \Delta x + x_C - x_F$. Thus we can bound $\Delta y_\text{final} \in [-h + m(\Delta x + x_C - x_F), h - m(\Delta x + x_C - x_F)]$ and $y_\text{final} \in [y_F -h + m\Delta x + m x_C - m x_F, y_F + h - m\Delta x - m x_C + m x_F])$. As above, we compare $y_\text{int}$ and $y_\text{final}$ and can see that $y_\text{int}$ and $y_\text{final}$ fall within the same intervals, with the former open and the latter closed. As above, we can state that any interior point of a polygon with $x_\text{int} > x_F$ lies inside the final polygon centered at $(x_F, y_F)$.

\subsection{Proof of Tightness}
\label{sec:tightness-proof}

Here we prove that $\text{safe}_\text{impl} \implies \text{safe}_\text{expl}$ by contraposition. Together with the proof in Sections 4.2-4.4, this completes our proof of equivalence of the implicit and explicit formulations.

As above, we assume the active corners $v_i, v_j$ do not change and we are in a rotated and shifted coordinate frame such that a line which runs through the active corners and the center of the polygon would be vertical. We also assume that we are in a segment of the trajectory with no active corner switches bounded by transition points, going from $(0,0)$ to $(x_F, y_F)$.

Let $(x_O, y_O)$ be the location of the obstacle. 
Let $(x_C, y_C)$ be the location of the center of the polygon. 
Let $(x_i, y_i)$ and $(x_j, y_j)$ be the locations of active corners $v_i$ and $v_j$. 
Let $\Delta x_i = x_i - x_C, \Delta y_i = y_i - y_C, \Delta x_j = x_j - x_C, \Delta y_j = y_j - y_C$ to represent the $x$ and $y$ distances from the center to an active corner. This implies $\Delta x_i = \Delta x_j = 0$ and that $x_C = x_i = x_j$, given the coordinate frame rotation. We will also assume that $y_i > y_j$, without loss of generality.

By definition, $\text{unsafe}_\text{impl}$ means $(x_O, y_O)$ is inside or on the border of the polygon at any point on the rotated trajectory from $(0,0)$ to $(x_F, y_F)$.
By definition, $\text{unsafe}_\text{expl}$ means either $$\left(y_O - f(x_O - \Delta x_i) - \Delta y_i\right) \cdot \left(y_O - f(x_O - \Delta x_j) - \Delta y_j\right) \leq 0 \\$$ when active corners $i$ and $j$ exist or $(x_O, y_O)$ is inside or on the border of the polygon when at transition points $(0,0)$ or $(x_F, y_F)$.

We prove that $\text{safe}_\text{impl} \implies \text{safe}_\text{expl}$ by proving the contrapositive $\text{unsafe}_\text{expl} \implies \text{unsafe}_\text{impl}$. We consider both cases for $(x_O, y_O)$ being in the explicit unsafe region.
\begin{itemize}
    \item  Case 1: There are no active corners, meaning $(x_O, y_O)$ is inside or on the border of the polygon at either $(0,0) $ or $(x_F, y_F)$. It follows trivially that $(x_O, y_O)$ is in the unsafe region by definition for the implicit formulation of the unsafe region. 
    \item Case 2: We have active corners $i$ and $j$, meaning $$\left(y_O - f(x_O) - \Delta y_i\right) \cdot \left(y_O - f(x_O) - \Delta y_j\right) \leq 0 \\ $$
\end{itemize}
We proceed with Case 2. Since we do have active corners, $x_O \in [0, x_F]$. Since our trajectory $f(x)$ is well-defined and continuous over the interval $[0, x_F]$, there exists a polygon centered on $(x_C, y_C) = (x_O, f(x_O))$. Given $x_O = x_C$, there are three possibilities for $y_O$, which we consider exhaustively.

\eat{
Note that we will either have 
$$\begin{cases}
  y > f(x_C) \text{ for } y > y_C \\
  y < f(x_C) \text{ for } y < y_C    
\end{cases}$$
or 
$$\begin{cases}
  y < f(x_C) \text{ for } y > y_C \\
  y > f(x_C) \text{ for } y < y_C    
\end{cases}$$

NOTE: need to justify why this is the case.

This also applies to $f(x_C, y - \Delta y_i)$ and $f(x_C, y - \Delta y_j)$, since they are the same function shifted horizontally. 
Without loss of generality, we assume that 
$\begin{cases}
  f(x_C,y) > 0 \text{ for } y > y_C  \\ 
  f(x_C,y) < 0 \text{ for } y < y_C   
\end{cases}$\\
}

\textbf{Case A: }$y_O > y_i$:

\indent \indent Let $y_h > y_i$. Then $y_h > y_j$ since we assume, without loss of generality, $y_i > y_j$. Rearrangement yields $y_h - y_i > 0$ and $y_h - y_j > 0$. Then $y_h - y_i + y_C - y_C> 0$ and $y_h - y_j + y_C - y_C > 0$. 

\indent \indent Therefore $y_h - (y_i - y_C) - f(x_C) = y_h - \Delta y_i - f(x_C) > 0$ and $y_h - (y_j - y_C) - f(x_C) = y_h - \Delta y_j - f(x_C) > 0$. This means $(y_h - f(x_C) - \Delta y_i)\cdot (y_h - f(x_C) - \Delta y_j) > 0$. 

\textbf{Case B: }$y_O < y_j$:

\indent \indent Let $y_l < y_j$. By the same reasoning as Case A, $(y_l - f(x_C) - \Delta y_i) \cdot (y_l - f(x_C) - \Delta y_j) > 0$, though the two terms multiplied are both negative.

\textbf{Case C: }$y_j \leq y_O \leq y_i$:

\indent \indent Let $y_j \leq y_m \leq y_i$. Then $y_m - y_j \geq 0$ and $y_m - y_i \leq 0$, so $y_m - y_j + y_C - y_C \geq 0$ and $y_m - y_i + y_C - y_C \leq 0$. Therefore $y_m - f(x_C) - \Delta y_i = y_m - f(x_C) - (y_i - y_C) \leq 0$ and $y_m - f(x_C) - \Delta y_j = y_m - f(x_C) - (y_j - y_C) \geq 0$. This means $$(y_m - f(x_C) - \Delta y_i)\cdot (y_m - f(x_C) - \Delta y_j) \leq 0$$

We have covered all possible values for $y_O$ when $x_O = x_C$ and we found that the only way for $(y_O - f(x_C) - \Delta y_i) \cdot (y_O - f(x_C) - \Delta y_j) \leq 0$ is to have $y_j \leq y_O \leq y_i$. Therefore $(y_O - f(x_C) - \Delta y_i) \cdot (y_O - f(x_C) - \Delta y_j) \leq 0\implies y_j \leq y_O \leq y_i$.

Since we are in a rotated coordinate system, $y_j, y_i$ refer to vertices of the polygon along a vertical centerline. Hence $(x_O, y_O)$ lies on that centerline, since we chose $x_C$ such that $x_C = x_O$. Thus $(x_O, y_O)$ is inside the polygon at the point $(x_C, y_C) = (x_O, f(x_O))$, which is the definition of $\text{unsafe}_\text{impl}$. Therefore $\text{unsafe}_\text{expl} \implies \text{unsafe}_\text{impl}$, or equivalently, $\text{safe}_\text{impl} \implies \text{safe}_\text{expl}$. Having shown both directions, we can state $\textbf{safe}_\text{impl} \mathbf{\iff} \textbf{safe}_\text{expl}$.

\clearpage
\newpage

\section{Applications}

\subsection{Dubins Path Example}
\label{appendix-dubins}

\begin{figure}[h]
    \centering
    \includegraphics[width=0.8\columnwidth]{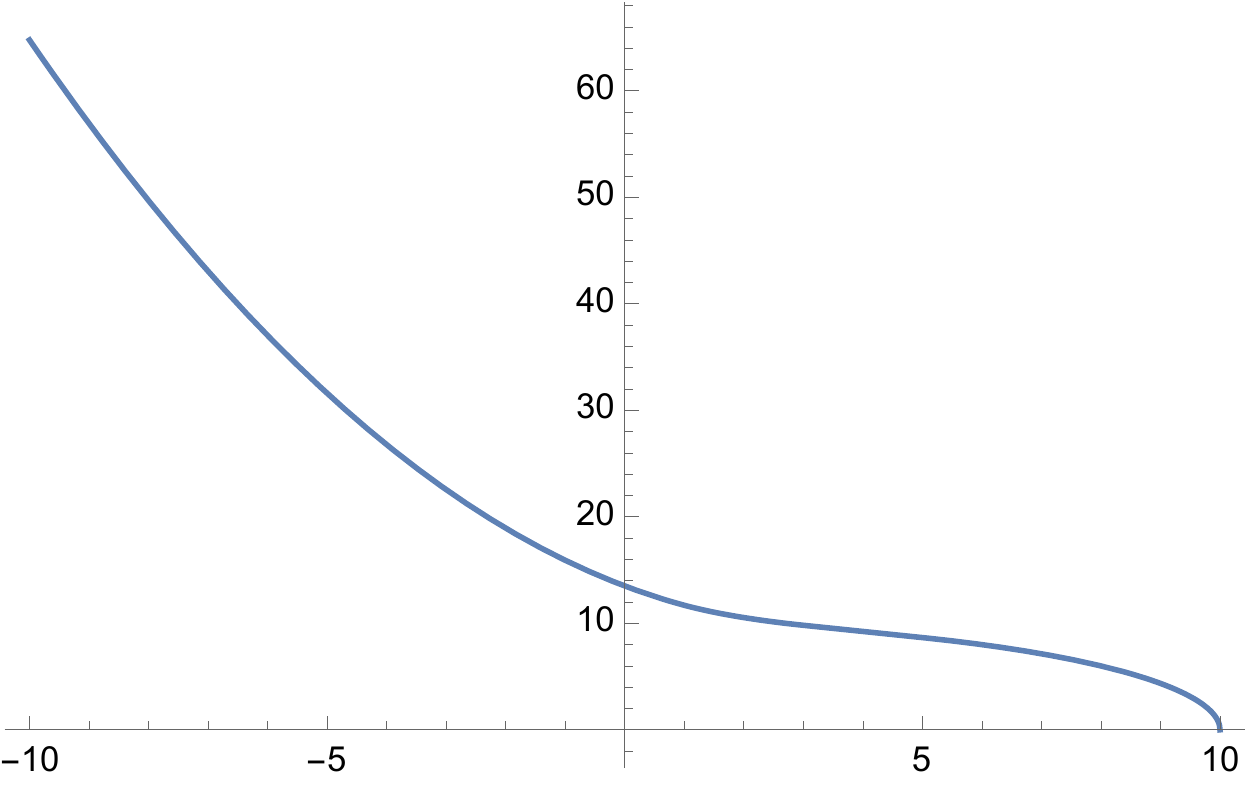}
    \caption{Dubins path instantiated with $R = 10, \theta = \frac{\pi}{3}, p=3, b=1$}
    \vspace{1cm}
    \label{fig:dubins-path-plot}
\end{figure}

The trajectory below in Equation (\ref{eqn:dubins-path}) is the equation for the Dubins path used in our evaluation above. Starting from the right at $x=R$, the path follows a circular path, then proceeds in a straight line after it has swept out angle $\theta$, then a straight line again until $x=p$, then another circular path until $x=b$, and a straight line after that. The entire path is $C^1$ smooth (function and its first derivative are continuous).
Because of its length and complexity, a fully symbolic explicit formulation for a rectangle with symbolic dimensions moving along this cannot be included in this paper but can be computed using our implementation available on GitHub.

\begin{strip}
\begin{equation}
    \begin{cases} \sqrt{R^{2} - x^{2}} & \text{for}\: x > \frac{R}{\sqrt{\tan^{2}{\left(\theta \right)} + 1}} \\R \sin{\left(\theta \right)} - \frac{- R \cos{\left(\theta \right)} + x}{\tan{\left(\theta \right)}} & \text{for}\: p < x \\\frac{R}{\sin{\left(\theta \right)}} - \frac{p}{\tan{\left(\theta \right)}} - \sqrt{\frac{p^{2}}{\cos^{2}{\left(\theta \right)}} - \left(- 2 p + x\right)^{2}} + \left|{p \tan{\left(\theta \right)}}\right| & \text{for}\: b < x \\\frac{R}{\sin{\left(\theta \right)}} - \frac{p}{\tan{\left(\theta \right)}} + \frac{\left(- b + x\right) \left(- 2 p + x\right) \left|{\cos{\left(\theta \right)}}\right|}{\sqrt{p^{2} - \left(b - 2 p\right)^{2} \cos^{2}{\left(\theta \right)}}} - \frac{\sqrt{- b^{2} \cos^{2}{\left(\theta \right)} + 4 b p \cos^{2}{\left(\theta \right)} - 4 p^{2} \cos^{2}{\left(\theta \right)} + p^{2}}}{\left|{\cos{\left(\theta \right)}}\right|} + \left|{p \tan{\left(\theta \right)}}\right| & \text{otherwise} \end{cases}
\end{equation}
\label{eqn:dubins-path}
\end{strip}

\subsection{Jeannin 2015}
\label{appendix-tacas}
Below is the explicit formulation of the unsafe region as plotted in Figure~\ref{fig:tacas-acas-paper-figure}. The following boolean formula is computed for a simplified version of the formulation in \cite{jeannin2015acastacas}, with a rectangular object of width 1.5 and height 1, and trajectory
\begin{equation}\begin{cases} c x^{2} & \text{for}\: b > x \\b^{2} c + 2 b c \left(- b + x\right) & \text{otherwise} \end{cases}\end{equation}

The boolean formula for the explicit formulation is below in Equation (\ref{eq:symbolic-acas-explicit}). It includes all the components from \eqref{safe-region-test-full}, such as testing if the obstacle is between the active corners as in \eqref{safe-region-test-corners-only}, clipping the function $f(\cdot)$ and using a function $g(\cdot)$ instead as in \eqref{def-g-x}, bounds ensuring tests apply only over a valid subregion as in Figure~\ref{fig:piecewise-traj-rules}, and testing for the notch at transition points and piecewise boundaries.

\begin{strip}
\begin{multline}
x \geq \frac{- b^{2} c w + b h + w y}{h} \wedge x \geq b - w \\
\wedge \left(- h + y - \begin{cases} b^{2} c & \text{for}\: b > - w + x \\b^{2} c + 2 b c \left(- b - w + x\right) & \text{for}\: w - x \geq -\infty \end{cases}\right) \left(h + y - \begin{cases} b^{2} c & \text{for}\: b > w + x \\b^{2} c + 2 b c \left(- b + w + x\right) & \text{for}\: w + x \leq \infty \end{cases}\right) \leq 0 \\ 
\vee x \geq \frac{b^{2} c w + b h - w y}{h} \wedge x \geq b - w \\ 
\wedge \left(- h + y - \begin{cases} b^{2} c & \text{for}\: b > w + x \\b^{2} c + 2 b c \left(- b + w + x\right) & \text{for}\: w + x \leq \infty \end{cases}\right) \left(h + y - \begin{cases} b^{2} c & \text{for}\: b > - w + x \\b^{2} c + 2 b c \left(- b - w + x\right) & \text{for}\: w - x \geq -\infty \end{cases}\right) \leq 0 \\ 
\vee x \leq \frac{- b^{2} c w + b h + w y}{h} \wedge x \leq b + w \\
\wedge \left(- h + y - \begin{cases} c \left(- w + x\right)^{2} & \text{for}\: b \geq - w + x \\b^{2} c & \text{otherwise} \end{cases}\right) \left(h + y - \begin{cases} c \left(w + x\right)^{2} & \text{for}\: b \geq w + x \\b^{2} c & \text{otherwise} \end{cases}\right) \leq 0 \vee \\
x \leq \frac{b^{2} c w + b h - w y}{h} \wedge x \leq b + w \\
\wedge \left(- h + y - \begin{cases} c \left(w + x\right)^{2} & \text{for}\: b \geq w + x \\b^{2} c & \text{otherwise} \end{cases}\right) \left(h + y - \begin{cases} c \left(- w + x\right)^{2} & \text{for}\: b \geq - w + x \\b^{2} c & \text{otherwise} \end{cases}\right) \leq 0  \\
\vee \left(- 2 h \left(- b - w + x\right) \geq 0 \wedge 2 h \left(- b + w + x\right) \geq 0 \wedge - 2 w \left(- b^{2} c - h + y\right) \geq 0 \wedge 2 w \left(- b^{2} c + h + y\right) \geq 0\right) \vee \\
\left(- 2 h \left(- b - w + x\right) \leq 0 \wedge 2 h \left(- b + w + x\right) \leq 0 \wedge - 2 w \left(- b^{2} c - h + y\right) \leq 0 \wedge 2 w \left(- b^{2} c + h + y\right) \leq 0\right)
\label{eq:symbolic-acas-explicit}
\end{multline}
\end{strip}

\clearpage
\newpage

\subsection{Adler 2019}
\label{appendix-uav}
Below is the explicit formulation of the unsafe region as plotted in Figure~\ref{fig:eytan-paper-figure}. The following boolean formula is computed for the formulation in \cite{adler2019formal}, with a regular hexagon of radius $2$ approximating a circular object and trajectory generated with parameters $R=10, \theta = \frac{\pi}{3}$:
\begin{equation}\begin{cases} \sqrt{100 - x^{2}} & \text{for}\: x > 5 \\- \frac{\sqrt{3} \left(x - 5\right)}{3} + 5 \sqrt{3} & \text{otherwise} \end{cases}\end{equation}

The fully symbolic trajectory for this application is 
\begin{equation}
    \begin{cases} \sqrt{R^{2} - x^{2}} & \text{for}\: x > \frac{R}{\sqrt{\tan^{2}{\left(\theta \right)} + 1}} \\R \sin{\left(\theta \right)} - \frac{- R \cos{\left(\theta \right)} + x}{\tan{\left(\theta \right)}} & \text{otherwise} \end{cases}
\end{equation}

The boolean formula for the explicit formulation is below in Equation (\ref{uav-explicit-numeric}). It includes all the components from \eqref{safe-region-test-full}, such as testing if the obstacle is between the active corners as in \eqref{safe-region-test-corners-only}, clipping the function $f(\cdot)$ and using a function $g(\cdot)$ instead as in \eqref{def-g-x}, bounds ensuring tests apply only over a valid subregion as in Figure~\ref{fig:piecewise-traj-rules}, and testing for the notch at transition points and piecewise boundaries. Because of its length and complexity, a fully symbolic explicit formulation cannot be included in this paper but can be computed using our implementation available on GitHub.

\begingroup
\allowdisplaybreaks
\begin{multline}
(x \geq -14 \ \wedge x \leq 7 \ \wedge  (\sqrt{3} x - y) (\sqrt{3} x - y + \frac{68 \sqrt{3}}{3}) \leq 0 \ \wedge \\  (y - \begin{cases} \frac{32 \sqrt{3}}{3} & \text{for}\: x - 1 < -12 \\- \frac{\sqrt{3} (x - 6)}{3} + 5 \sqrt{3} & \text{for}\: x - 1 \leq 5 \\5 \sqrt{3} & \text{otherwise} \end{cases} - \sqrt{3}) \times \\ (y - \begin{cases} \frac{32 \sqrt{3}}{3} & \text{for}\: x + 1 < -12 \\- \frac{\sqrt{3} (x - 4)}{3} + 5 \sqrt{3} & \text{for}\: x + 1 \leq 5 \\5 \sqrt{3} & \text{otherwise} \end{cases} + \sqrt{3}) \leq 0) \\ \ \vee   (x \geq 3 \ \wedge \\  x \leq 2 + 5 \sqrt{3} \ \wedge \\  (\sqrt{3} x - y) (\sqrt{3} x - y - 10) \leq 0 \ \wedge \\  (y - \begin{cases} 5 \sqrt{3} & \text{for}\: x - 1 < 5 \\\sqrt{100 - (x - 1)^{2}} & \text{for}\: x - 1 \leq 5 \sqrt{3} \\5 & \text{otherwise} \end{cases} - \sqrt{3}) \times \\ (y - \begin{cases} 5 \sqrt{3} & \text{for}\: x + 1 < 5 \\\sqrt{100 - (x + 1)^{2}} & \text{for}\: x + 1 \leq 5 \sqrt{3} \\5 & \text{otherwise} \end{cases} + \sqrt{3}) \leq 0) \ \vee \\  (x \geq -2 + 5 \sqrt{3} \ \wedge \\  x \leq 12 \ \wedge \\  y (y - 5) \leq 0 \ \wedge \\  (y - \begin{cases} 5 & \text{for}\: x - 2 < 5 \sqrt{3} \\\sqrt{100 - (x - 2)^{2}} & \text{for}\: x - 2 \leq 10 \\0 & \text{otherwise} \end{cases}) (y - \begin{cases} 5 & \text{for}\: x + 2 < 5 \sqrt{3} \\\sqrt{100 - (x + 2)^{2}} & \text{for}\: x + 2 \leq 10 \\0 & \text{otherwise} \end{cases}) \leq 0) \ \vee \\  (- 2 y + 2 \sqrt{3} \geq 0 \ \wedge  - y - \sqrt{3} (x - 12) \geq 0 \ \wedge  y + \sqrt{3} (x - 8) \geq 0 \ \wedge \\  2 y + 2 \sqrt{3} \geq 0 \ \wedge  - y + \sqrt{3} (x - 9) + \sqrt{3} \geq 0 \ \wedge \\  y - \sqrt{3} (x - 11) + \sqrt{3} \geq 0) \ \vee  (- 2 y + 12 \sqrt{3} \geq 0 \ \wedge \\  2 y - 8 \sqrt{3} \geq 0 \ \wedge  - y - \sqrt{3} (x - 7) + 5 \sqrt{3} \geq 0 \ \wedge \\  - y + \sqrt{3} (x - 4) + 6 \sqrt{3} \geq 0 \\ \wedge  y - \sqrt{3} (x - 6) - 4 \sqrt{3} \geq 0 \ \wedge \\  y + \sqrt{3} (x - 3) - 5 \sqrt{3} \geq 0) \ \vee  (- 2 y + \frac{70 \sqrt{3}}{3} \geq 0 \ \wedge \\  2 y - \frac{58 \sqrt{3}}{3} \geq 0 \ \wedge  - y - \sqrt{3} (x + 10) + \frac{32 \sqrt{3}}{3} \geq 0 \ \wedge  - y + \sqrt{3} (x + 13) + \\ \frac{35 \sqrt{3}}{3} \geq 0 \ \wedge  y - \sqrt{3} (x + 11) - \frac{29 \sqrt{3}}{3} \geq 0 \ \wedge \\  y + \sqrt{3} (x + 14) - \frac{32 \sqrt{3}}{3} \geq 0) \ \vee  (- 2 y + 2 \sqrt{3} + 10 \geq 0 \ \wedge \\  - y - \sqrt{3} (x - 5 \sqrt{3} - 2) + 5 \geq 0 \ \wedge \\  y + \sqrt{3} (x - 5 \sqrt{3} + 2) - 5 \geq 0 \ \wedge  2 y - 10 + 2 \sqrt{3} \geq 0 \ \wedge \\  - y + \sqrt{3} (x - 5 \sqrt{3} + 1) + \sqrt{3} + 5 \geq 0 \ \wedge  y - \sqrt{3} (x - 5 \sqrt{3} - 1) - 5 + \sqrt{3} \geq 0) \ \vee \\  (- 2 y + 2 \sqrt{3} \leq 0 \ \wedge  - y - \sqrt{3} (x - 12) \leq 0 \ \wedge  y + \sqrt{3} (x - 8) \leq 0 \\ \wedge  2 y + 2 \sqrt{3} \leq 0 \ \\ \wedge   - y + \sqrt{3} (x - 9) + \sqrt{3} \leq 0 \ \wedge  y - \sqrt{3} (x - 11) + \sqrt{3} \leq 0) \ \vee \\  (- 2 y + 12 \sqrt{3} \leq 0 \ \wedge  2 y - 8 \sqrt{3} \leq 0 \ \wedge \\  - y - \sqrt{3} (x - 7) + 5 \sqrt{3} \leq 0 \ \wedge  - y + \sqrt{3} (x - 4) + 6 \sqrt{3} \leq 0 \ \wedge \\  y - \sqrt{3} (x - 6) - 4 \sqrt{3} \leq 0 \  \\ \wedge  y + \sqrt{3} (x - 3) - 5 \sqrt{3} \leq 0) \ \vee   (- 2 y + \frac{70 \sqrt{3}}{3} \leq 0 \ \wedge \\  2 y - \frac{58 \sqrt{3}}{3} \leq 0 \ \wedge   - y - \sqrt{3} (x + 10) + \frac{32 \sqrt{3}}{3} \leq 0 \ \wedge \\  - y + \sqrt{3} (x + 13) + \\ \frac{35 \sqrt{3}}{3} \leq 0 \ \wedge  y - \sqrt{3} (x + 11) - \frac{29 \sqrt{3}}{3} \leq 0 \ \wedge \\  y + \sqrt{3} (x + 14) - \frac{32 \sqrt{3}}{3} \leq 0) \ \vee \\  (- 2 y + 2 \sqrt{3} + 10 \leq 0 \ \wedge  - y - \sqrt{3} (x - 5 \sqrt{3} - 2) + 5 \leq 0 \ \\ \wedge  y + \sqrt{3} (x - 5 \sqrt{3} + 2) - 5 \leq 0 \ \wedge  \\ 2 y - 10 + 2 \sqrt{3} \leq 0 \ \wedge  - y + \sqrt{3} (x - 5 \sqrt{3} + 1) + \sqrt{3} + 5 \leq 0 \ \wedge \\  y - \sqrt{3} (x - 5 \sqrt{3} - 1) - 5 + \sqrt{3} \leq 0)
\label{uav-explicit-numeric}
\end{multline}

\endgroup

\end{document}